\documentclass{article}
\usepackage{spconf}
\usepackage{amsmath,amsfonts}
\usepackage{algorithmic}
\usepackage{algorithm}
\usepackage{array}
\usepackage[caption=false,font=normalsize,labelfont=sf,textfont=sf]{subfig}
\usepackage{textcomp}
\usepackage{stfloats}
\usepackage{url}
\usepackage{verbatim}
\usepackage{graphicx}
\usepackage{cite}
\usepackage{picture}
\usepackage{verbatim}
\usepackage{pgfplots}
\usepackage{booktabs}
\usepackage{multicol}
\usepackage{multirow}
\usepackage{xcolor}
\usepackage{amssymb}
\usepackage{enumitem}

\newcommand\etal{\emph{et al. }}
\newcommand\red{\textcolor{red}}
\newcommand\blue{\textcolor{blue}}

\title{Multi-dimension Queried and Interacting Network for Stereo Image Deraining}

\begin{document}
%

\name{Yuanbo Wen$^{1}$ \qquad Tao Gao$^{1*}$ \qquad Ziqi Li$^{1}$ \qquad Jing Zhang$^{2}$ \qquad Ting Chen$^{1*}$}
\address{$^{1}$ School of Information Engineering, Chang'an University, Xi'an, China; $^{2}$ College of Engineering
\\ and Computer Science, Australian National University, Canberra, ACT, Australia}

\maketitle
\begin{abstract}
Eliminating the rain degradation in stereo images poses a formidable challenge, which necessitates the efficient exploitation of mutual information present between the dual views.
To this end, we devise MQINet, which employs multi-dimension queries and interactions for stereo image deraining.
More specifically, our approach incorporates a context-aware dimension-wise queried block (CDQB). This module leverages dimension-wise queries that are independent of the input features and employs global context-aware attention (GCA) to capture essential features while avoiding the entanglement of redundant or irrelevant information.
Meanwhile, we introduce an intra-view physics-aware attention (IPA) based on the inverse physical model of rainy images. IPA extracts shallow features that are sensitive to the physics of rain degradation, facilitating the reduction of rain-related artifacts during the early learning period.
Furthermore, we integrate a cross-view multi-dimension interacting attention mechanism (CMIA) to foster comprehensive feature interaction between the two views across multiple dimensions.
Extensive experimental evaluations demonstrate the superiority of our model over EPRRNet and StereoIRR, achieving respective improvements of 4.18 dB and 0.45 dB in PSNR.
Code and models are available at \url{https://github.com/chdwyb/MQINet}.

\end{abstract}
\section{Introduction}
\label{sec:intro}

\footnotetext[1]{Corresponding author: Tao Gao, Ting Chen}
\footnotetext[2]{This research was partially supported by the National Key R \& D Program of China under Grants 2019YFE0108300, the National Natural Science Foundation of China under Grants 52172379, 62001058 and U1864204, the Fundamental Research Funds for the Central University under Grants 300102242901.}

As stereo image processing techniques continue to advance, there has been a growing focus on stereo image deraining. This development is driven by the recognition of the complementary information available in left- and right-view images and its substantial impact on the performance of subsequent high-level vision algorithms \cite{gao2023frequency, zhang2022single, yang2021fast, ren2023semi, gao2023towards}.
For instance, Zhang \etal \cite{zhang2020beyond, zhang2022beyond} introduce a parallel network that leverages both stereo images and semantic information to effectively remove rain streaks from stereo images.
Subsequently, Wei \etal \cite{wei2022stereo} devise a dual-view mutual attention mechanism to facilitate interaction between distinct views in the context of stereo image deraining.
Recently, Nie \etal \cite{nie2023context} introduce a network that focuses on context and detail interaction to eliminate both rain streaks and raindrops in stereo images.
Nevertheless, existing methods fall short in extracting information from multiple dimensions, resulting in limited performance.

In this work, we propose a multi-dimensional queried and interacting network for stereo image deraining, named MQINet.
Specifically, we introduce a novel context-aware dimension-wise queried block (CDQB) as the fundamental learning unit.
Initially, we partition the input features into four distinct segments, each of which undergoes unique processing techniques \cite{chen2023run, ruan2023ege}.
CDQB employs dimension-wise queries on the $hw$, $ch$, and $cw$ dimensions to interrogate their respective quarter components.
By utilizing these queries, we manage the information flow into the feed-forward network without explicitly encoding the content of the input features, which results in enhanced generalization capabilities and a reduction in irrelevant information.
For the remaining quarter components, we leverage global context-aware attention (GCA) to extract global contextual information and derive partially self-relevant representations.
Our CDQB is better equipped to capture essential patterns and features while avoiding entanglement in redundant or irrelevant information.

Additionally, we introduce an intra-view physics-aware attention (IPA) mechanism grounded in the inverse physical model of rainy images.
Existing methods \cite{zhang2020beyond, wei2022stereo, nie2023context} for stereo image deraining predominantly involve mapping rain-distorted images to high-dimensional feature spaces using single-layer convolutions, overlooking the deeper exploration of the inherent characteristics within the composition of rain streaks and the pristine background.
We employ various learning operations to approximate the corresponding components within the inverse physical model (Eq. \ref{eq:physics}), thereby enhancing the accuracy of shallow feature extraction.
Our IPA framework facilitates the attenuation of rain-induced degradation during the early stages of learning, leading to substantial improvements in overall performance for stereo image deraining.

Moreover, existing mutual modules \cite{chu2022nafssr, wei2022stereo} in stereo image processing methods primarily focus on feature interaction in the $w$ dimension, which proves insufficient for effectively extracting interrelated features within the $h$ and $c$ dimensions.
To this end, we introduce a cross-view multi-dimension interacting attention mechanism (CMIA) that facilitates dual-view learning across all three dimensions, resulting in a superior mutual feature representation.
Specifically, we partition the query and value representations of input features from the left and right views into three distinct segments, each corresponding to a different dimension.
For instance, the mutual attention maps in the $h$ dimension are computed using $q_{l, h}$ and $q_{r, h}$.
Subsequently, the value representations of both views, denoted as $v_{l, h}$ and $v_{r, h}$, are enhanced using these mutual attention maps.
Enabling feature interaction across multiple dimensions promotes more comprehensive information exchange between the two views, consequently enhancing the overall performance of rain removal in single-view rain-degraded images.

\begin{figure}
\vspace{-2mm}
    \centering
    \includegraphics[width=\linewidth]{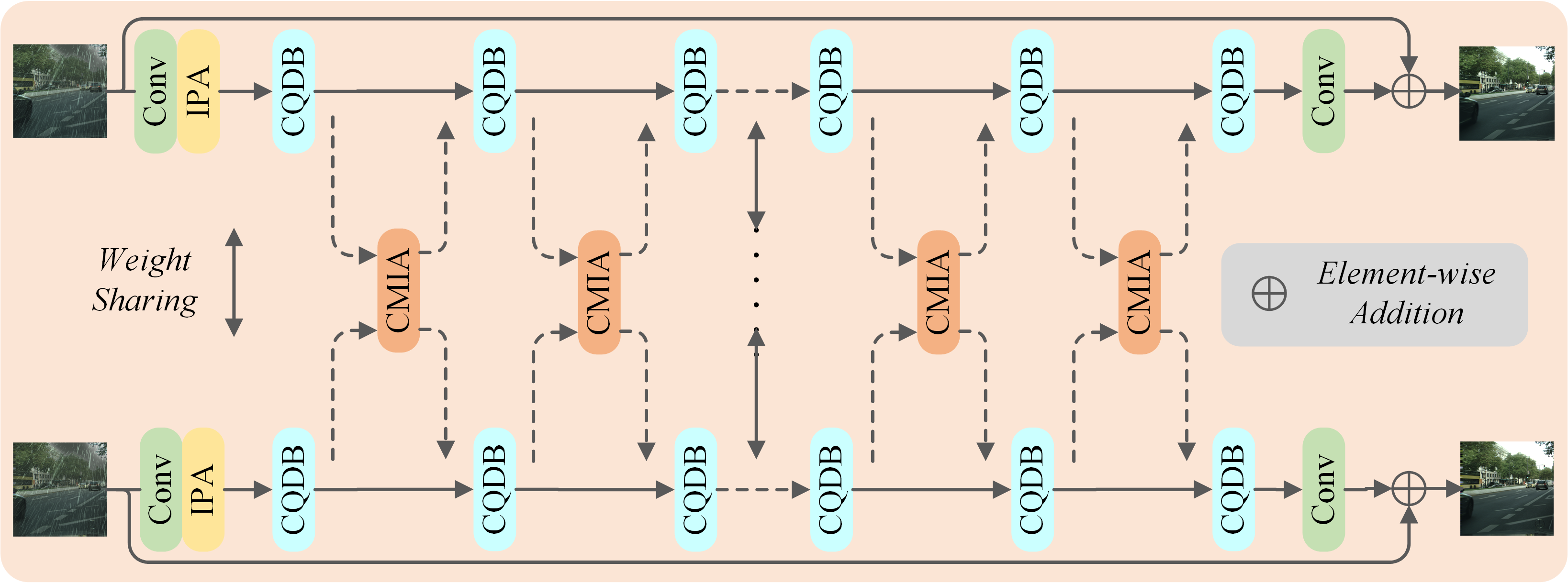}
    \caption{Architectural overview of our proposed multi-dimension queried and interacting network for stereo image deraining.}
    \label{fig:network}
    \vspace{-2mm}
\end{figure}

Figure \ref{fig:network} illustrates the overall pipeline of our proposed multi-dimension queried and interacting network.
Our contributions can be summarized as follows:
1) We propose a context-aware dimension-wise queried block that utilizes the input-independent dimension-wise queries and context-aware attention to
capture essential features without getting entangled in redundant or irrelevant information.
2) Based on the inverse physical model of rainy images, we devise an intra-view physics-aware attention to extract the physics-aware shallow features, facilitating the physics-aware feature learning of rain degradation in early period.
3) we introduce a cross-view multi-dimension interacting attention that achieves the comprehensive feature interaction of two views from multiple dimensions.
4) Extensive experimental results verify the effectiveness of our proposed method.

\section{Proposed Method}

\subsection{Context-aware Dimension-wise Queried Block}

\begin{figure}
\vspace{-2mm}
    \centering
    \includegraphics[width=\linewidth]{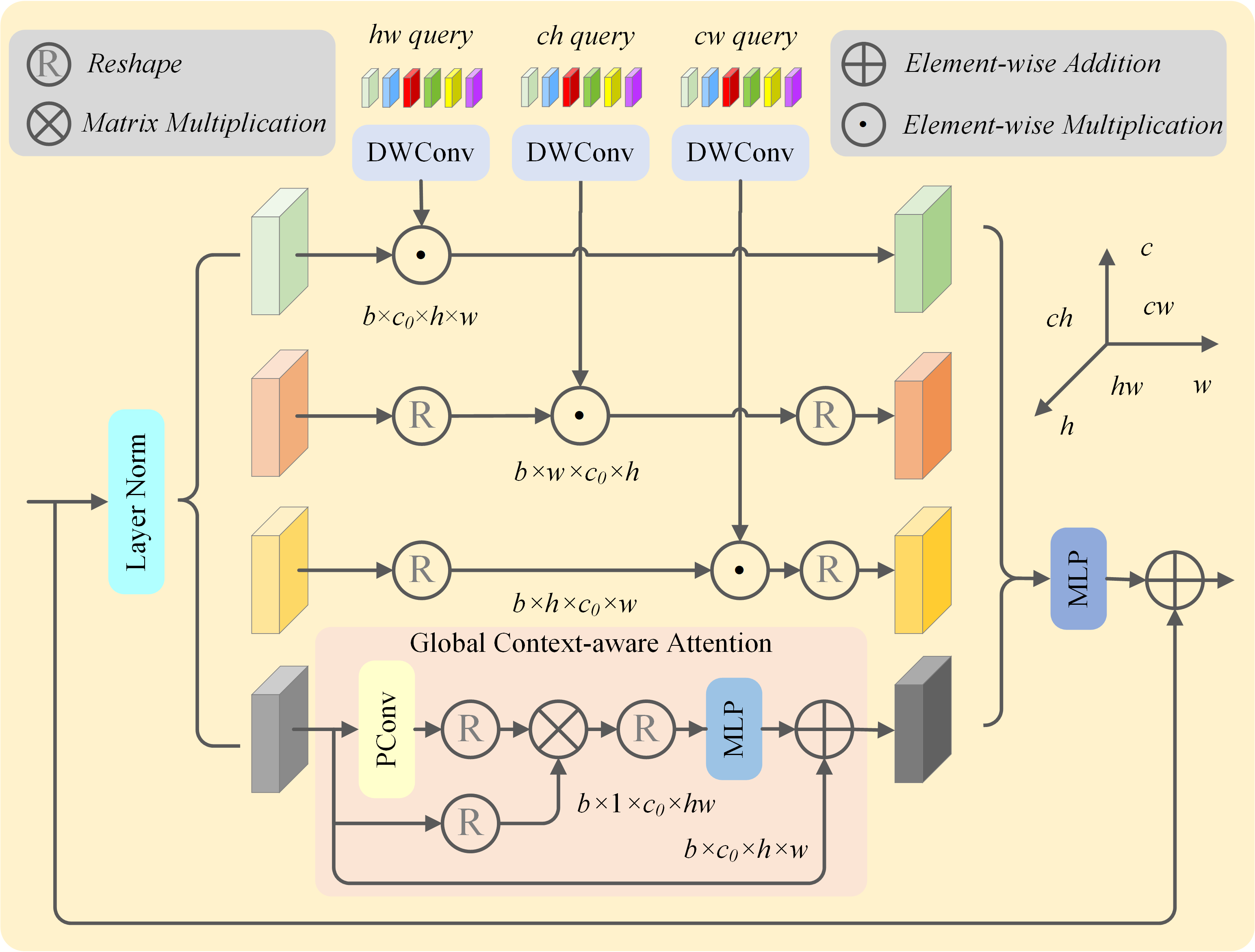}
    \caption{Illustration of our proposed context-aware dimension-wise queried block (CDQB). The information flow is controlled by three dimension-wise queries and the global context-aware attention (GCA).}
    \label{fig:cdqb}
    \vspace{-2mm}
\end{figure}

In this block, we discern the three quarter of input features along the dimensions of channel ($c$), height ($h$) and width ($w$). For each feature group, we apply learnable queries on the $hw$, $ch$, and $cw$ planes, respectively.
A distinctive characteristic of our approach is the refusal to directly learn the content of input features.
Instead, we employ three groups of dimension-wise queries to conduct partial feature learning and control the informative features flow into the feed-forward network.
Given an input feature matrix $\textbf{X}$ and randomly initialized learnable queries $\textbf{Q}$, we resize $\textbf{Q}$ to match the size of $\textbf{X}$ using bi-linear interpolation.
Meanwhile, we utilize depth-wise convolution to enhance the spatial representation of these dimension-wise queries.
We segregate the features into different dimensions, $\textbf{X}_{hw}\in \mathbb{R}^{c\times h\times w}$, $\textbf{X}_{ch}\in \mathbb{R}^{w\times c\times h}$, $\textbf{X}_{cw}\in \mathbb{R}^{h\times c\times w}$, and $\textbf{X}_{0}$ representing the remaining quarter of features.
We multiply the corresponding dimension-wise queries $\textbf{Q}_{hw}\in \mathbb{R}^{c\times h\times w}$, $\textbf{Q}_{ch}\in \mathbb{R}^{w\times c\times h}$, and $\textbf{Q}_{cw}\in \mathbb{R}^{h\times c\times w}$ with the respective feature channels.
For $\textbf{X}_{0}$, we utilize global context-aware attention (GCA) to extract the global context information to achieve partial self-relevant feature representation.
Finally, we concatenate the four groups of enhanced features and employ a simple feed-forward network to perform feature transformation.
Therefore, the multi-axis partial queried features are derived from
\begin{equation}
    \begin{aligned}
        & \left[\textbf{Q}_{hw}, \textbf{Q}_{ch}, \textbf{Q}_{cw}\right] = DWC(\textbf{Q}), \\
        & \textbf{X}_{queried}^{'}=Concat(\{\textbf{Q}_{ij}\odot\textbf{X}_{ij}\}, GCA(\textbf{X}_{0})), \\
    \end{aligned}
\end{equation}
where $DWC(\cdot)$ is the depth-wise convolution, $Concat(\cdot)$ indicates the concatenation operation, and $ij\in \{hw, ch, cw\}$.

The utilization of queries serves a twofold purpose.
Firstly, it significantly reduces the computational complexity, making it more efficient and suitable for basic learning block in remote sensing image dehazing.
Secondly, the introduction of dimension-wise queries enhances the generalization performance as demonstrated in the works of \cite{valanarasu2022transweather, ruan2023ege, gao2023frequency}.
By refraining from explicitly learning strictly related information of the input feature itself, our CDQB is better equipped to capture essential patterns and features without getting entangled in redundant or irrelevant information.

\subsection{Intra-view Physics-aware Attention}

\begin{figure}
\vspace{-2mm}
    \centering
    \includegraphics[width=\linewidth]{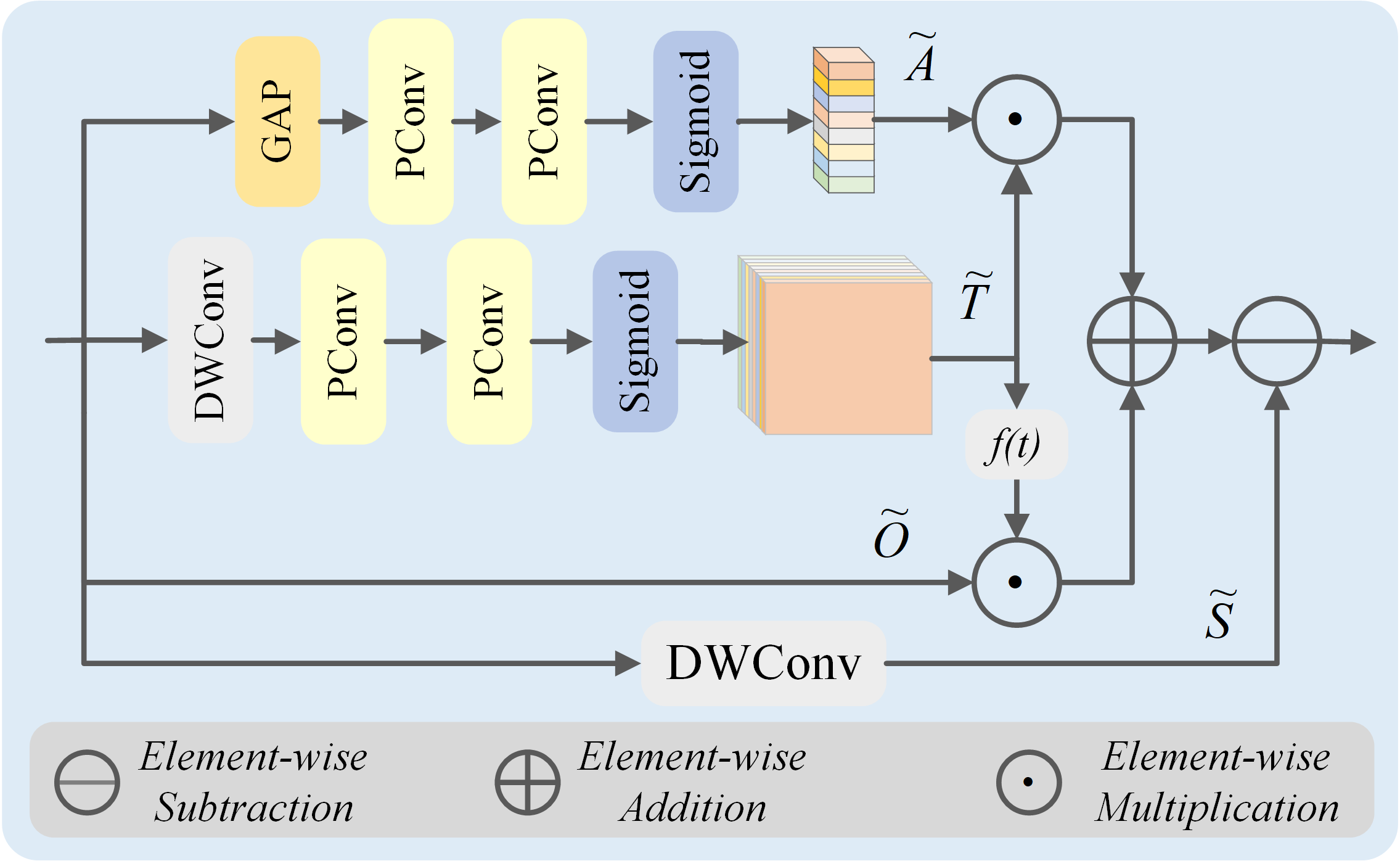}
    \caption{Illustration of our proposed intra-view physics-aware attention (IPA). It extracts the shallow features of the rain-degraded images based on the inverse physical model.}
    \label{fig:ipa}
    \vspace{-2mm}
\end{figure}

Current methods \cite{zhang2020beyond, wei2022stereo, zhang2022beyond, nie2023context} for stereo image deraining directly extract the shallow features with one-layer convolution, which ignores the in-depth characteristics of degradation composition, leading to an un-accurate feature representation of the rain images.
To this end, we propose an intra-view physics-aware attention (IPA), which generates the mapping features based on the inverse physical model.
Specifically, according to \cite{fu2017clearing, yang2020single, zhang2023data}, rainy image is formulated as
\begin{equation}
    O=\alpha (B + S) + (1- \alpha) A,
\end{equation}
where $O$, $B$, $S$, $\alpha$ and $A$ denote the rainy image, clean background, rain layers, transmission map and global atmospheric light.
Therefore, the clean background $B$ can be obtained by
\begin{equation}
    \textbf{B}=\textbf{T}\odot\textbf{O}-\textbf{S}+(\textbf{1}-\textbf{T})\odot\textbf{A},
\end{equation}
where $\textbf{T}$ denotes $\frac{1}{\alpha}$, $\odot$ is the element-wise multiplication, $\textbf{B}$, $\textbf{O}$, $\textbf{S}$ and $\textbf{A}$ are the matrix representation of $B$, $O$, $S$ and $A$, respectively.
We define the network parameters as $\textbf{K}$, then we can obtain the shallow features by
\begin{equation}
    \textbf{K}\circledast\textbf{B}=\textbf{K}\circledast\textbf{T}\odot\textbf{K}\circledast\textbf{O}-\textbf{K}\circledast\textbf{S}+\textbf{K}\circledast(\textbf{1}-\textbf{T})\odot\textbf{K}\circledast\textbf{A},
\end{equation}
where $\circledast$ denotes the convolutional operation.
We define $\tilde{\textbf{O}}$, $\tilde{\textbf{T}}$, $\tilde{\textbf{S}}$, $\tilde{\textbf{A}}$ as the approximation of $\textbf{K}\otimes\textbf{O}$, $\textbf{K}\otimes\textbf{T}$, $\textbf{K}\otimes\textbf{S}$ and $\textbf{K}\otimes\textbf{A}$. Therefore, we can obtain the physics-aware features by
\begin{equation}
\label{eq:physics}
    \tilde{\textbf{B}}=\tilde{\textbf{T}}\odot\tilde{\textbf{O}}-\tilde{\textbf{S}}+(\textbf{1}-\tilde{\textbf{T}})\odot\tilde{\textbf{A}},
\end{equation}
As Figure \ref{fig:ipa} shown, we obtain $\tilde{\textbf{A}}$ and $\tilde{\textbf{T}}$ by
\begin{equation}
\begin{aligned}
    & \tilde{\textbf{A}} & =\sigma(MLP(GAP(\textbf{X}))), \\
    & \tilde{\textbf{T}} & =\sigma(MLP(DWC(\textbf{X}))),
\end{aligned}
\end{equation}
where $MLP$ denotes the feed-forward network, $GAP$ indicates the global average pooling. The $\tilde{\textbf{S}}$ is obtained by a depth-wise convolution, then the physics-aware features are generated by Eq. \ref{eq:physics}.

\subsection{Cross-view Multi-dimension Interacting Attention}

\begin{figure}
\vspace{-2mm}
    \centering
    \includegraphics[width=\linewidth]{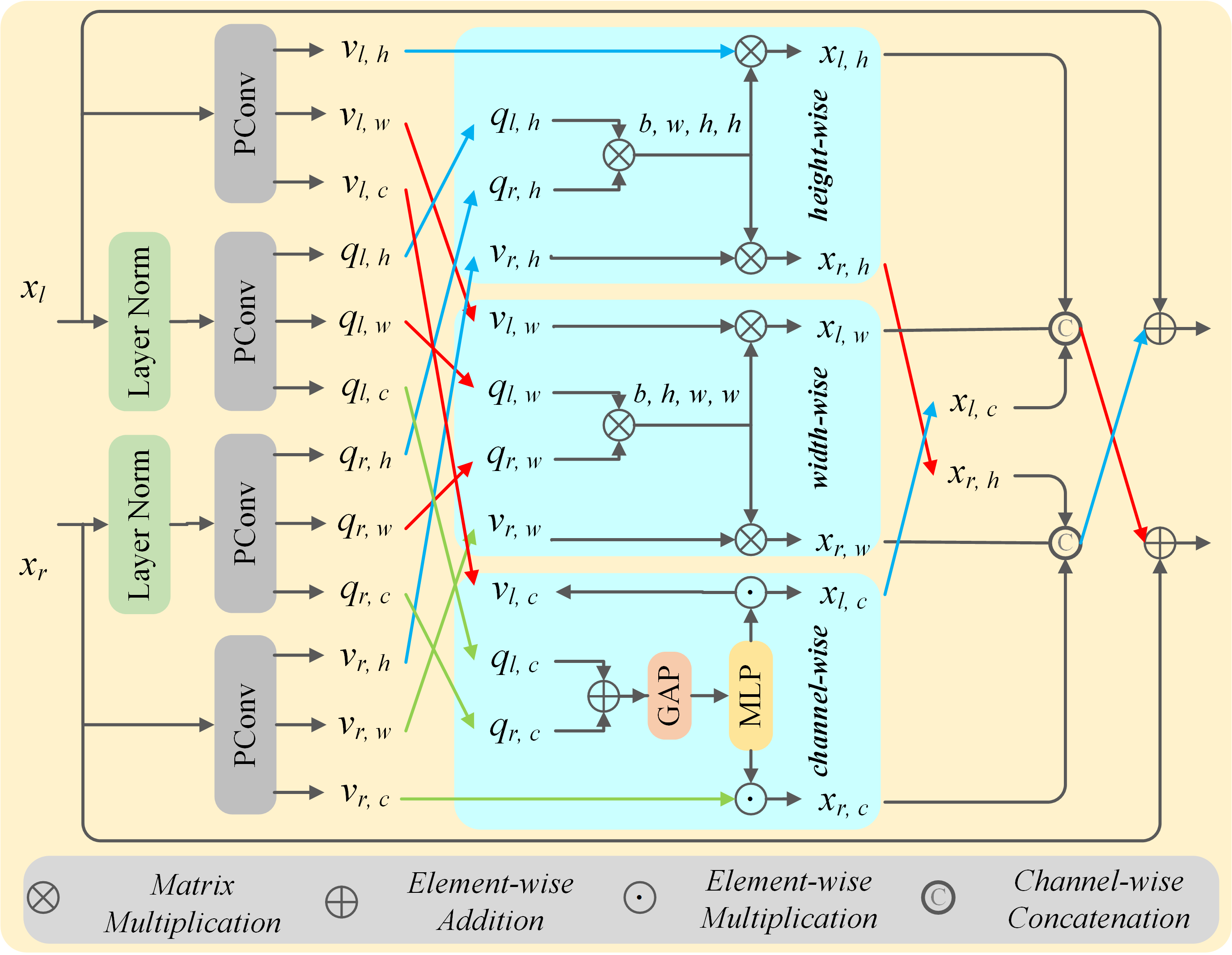}
    \caption{Illustration of our proposed cross-view multi-dimension interacting attention (CMIA). It conducts dual-view interaction from three dimensions and improves the mutual information communication.}
    \label{fig:cmia}
    \vspace{-2mm}
\end{figure}

Current stereo image processing methodologies \cite{chu2022nafssr, wei2022stereo, zhang2020beyond} exclusively focus on feature interaction along the $w$ dimension, neglecting the efficient utilization of mutual information within the $c$ and $h$ dimensions.
To this end, we introduce a cross-view multi-dimension interacting attention (CMIA) mechanism.
Specifically, we partition the input features into three segments, each aligns with a distinct dimension-wise representation.
For the first part, we achieve $h$-wise interaction by
\begin{equation}
\begin{aligned}
    & x_{l, h} & =\sigma(q_{l, h}\otimes q_{r, h}) \otimes v_{l, h}, \\
    & x_{r, h} & =\sigma(q_{l, h}\otimes q_{r, h})^{t} \otimes v_{r, h},
\end{aligned}
\end{equation}
where $t$ denotes the matrix transposing operation, $x_{l, h}$ is the partial features from the left-view height-wise representation.
Likewise, the $w$-wise interaction can be formulated by
\begin{equation}
\begin{aligned}
    & x_{l, w} & =\sigma(q_{l, w}\otimes q_{r, w}) \otimes v_{l, w}, \\
    & x_{r, w} & =\sigma(q_{l, w}\otimes q_{r, w})^{t} \otimes v_{r, w}.
\end{aligned}
\end{equation}
As for the $w$-wise interaction, we conduct selective kernel enhancement \cite{li2019selective, song2023vision} as
\begin{equation}
\begin{aligned}
    & x_{l, c} & = MLP(GAP(q_{l, c} + q_{r, c})) \odot v_{l, c}, \\
    & x_{r, c} & = MLP(GAP(q_{l, c} + q_{r, c})) \odot v_{r, c}, \\
\end{aligned}
\end{equation}
Consequently, we concatenate the multi-dimension mutual features to obtain the cross-view interacting features.
\begin{equation}
    \begin{aligned}
        & x_l^{i} & = x_l^{i-1} + Concat(x_{r, h}, x_{r, w}, x_{r, c}), \\
        & x_r^{i} & = x_r^{i-1} + Concat(x_{l, h}, x_{l, w}, x_{l, c}),
    \end{aligned}
\end{equation}
where $x^{i-1}$ and $x^{i}$ denote the input and output features.
CMIA conducts bidirectional information interaction across various dimensions, effectively harnessing additional-view features to effectively remove rain streaks from the rainy images in the current view.

\section{Experiments}

\subsection{Implementation Specifications}
We conduct experiments on NVIDIA Tesla A100 GPU with total 200 epoch, while the batch and patch sizes are initially set to 32 and 128, which are changed to 16 and 256 when epoch is 100.
We utilize the Adam optimizer \cite{kingma2014adam} with an initial learning rate of 0.01, which is gradually reduced to $1\times 10^{-7}$ by cosine annealing decay \cite{loshchilov2016sgdr}.

\subsection{Datasets}

RainKITTI2012 \cite{zhang2022beyond, geiger2013vision} dataset, comprising 4062 stereo image pairs for training and 4085 stereo image pairs for testing.
RainKITTI2015 dataset \cite{zhang2022beyond, geiger2013vision}, which includes 4200 training and 4189 testing stereo image pairs.
StereoCityscapes dataset \cite{wei2022stereo, cordts2016cityscapes}, with 2975 stereo image pairs in the training subset and 1525 stereo image pairs in the testing subset.

\subsection{Quantitative and Visual Comparisons}

\begin{figure}
    \centering
    \includegraphics[width=1.6cm]{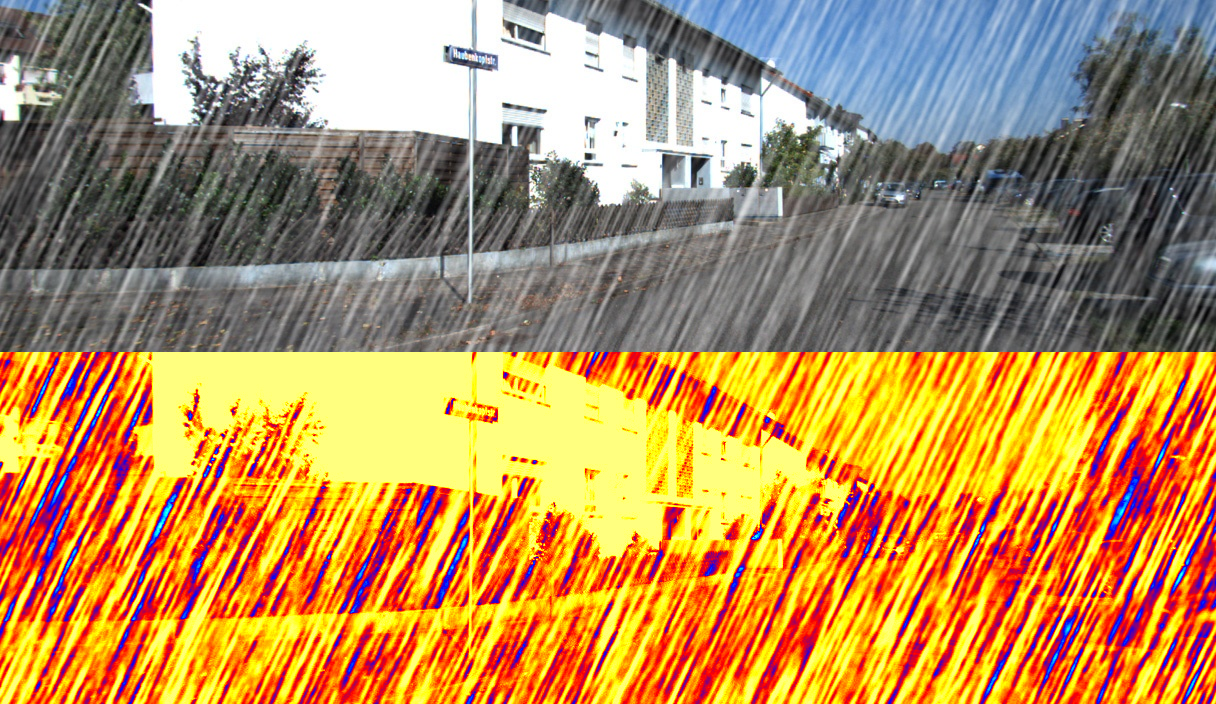}
    \includegraphics[width=1.6cm]{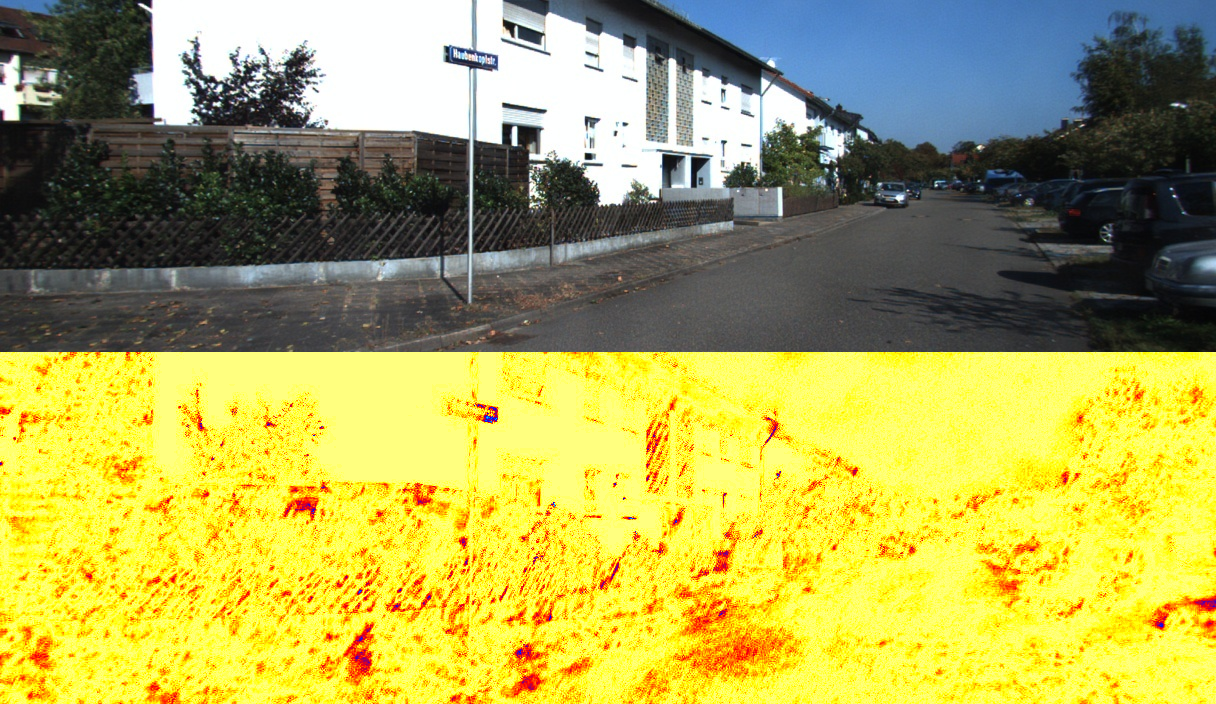}
    \includegraphics[width=1.6cm]{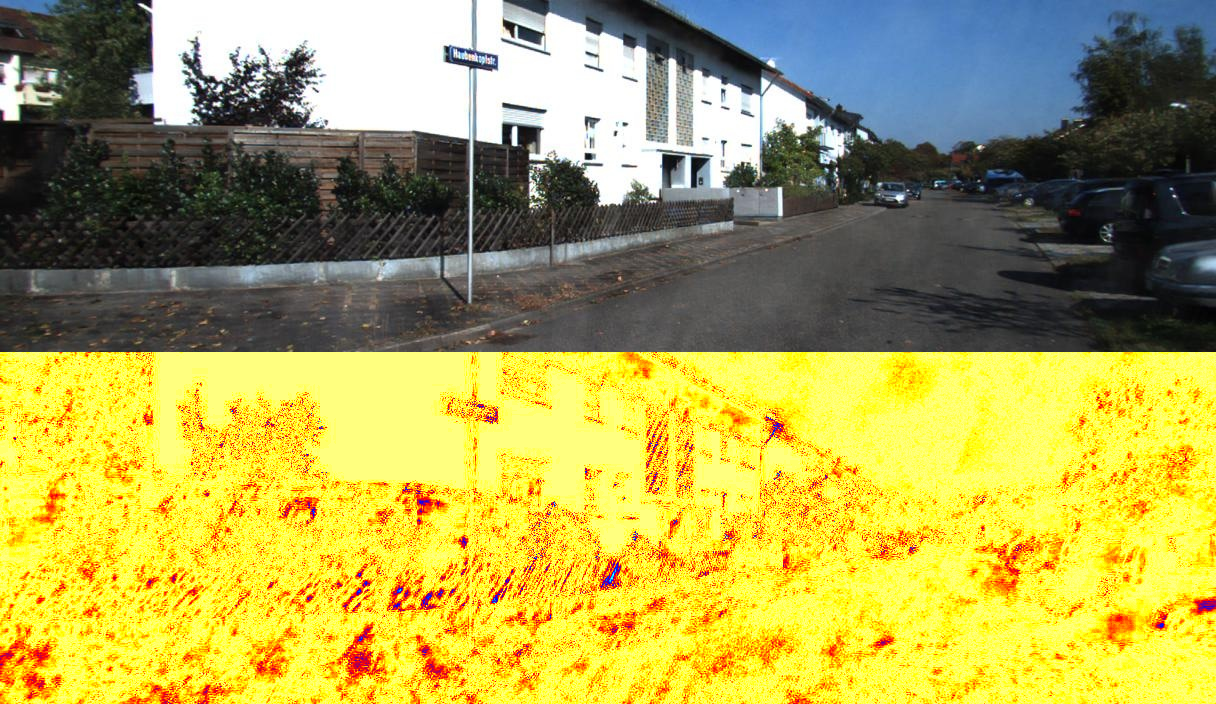}
    \includegraphics[width=1.6cm]{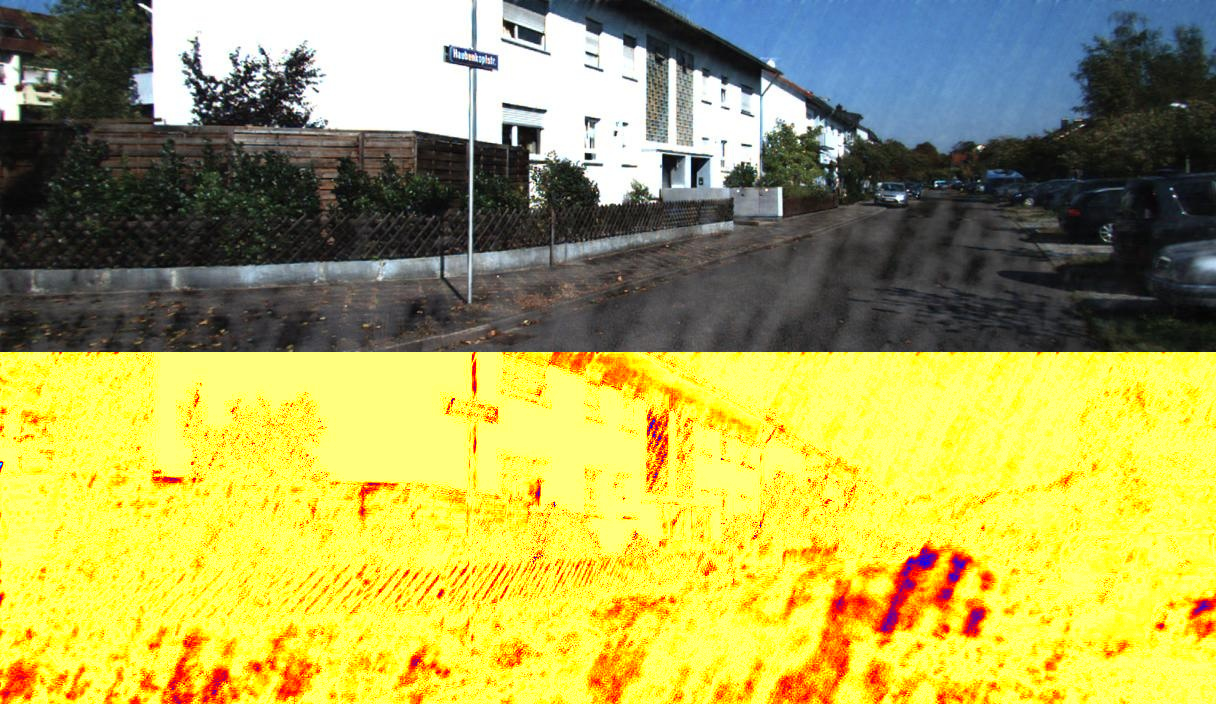}
    \includegraphics[width=1.6cm]{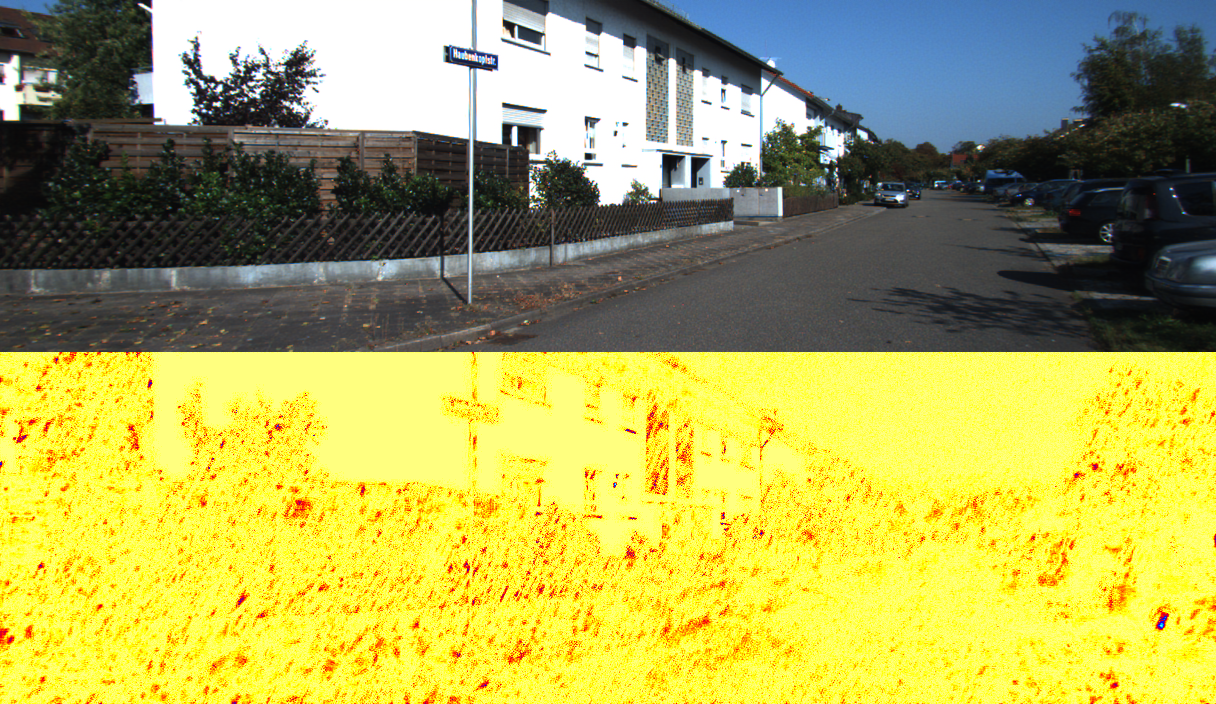}

    \vspace{0.5mm}
    \includegraphics[width=1.6cm]{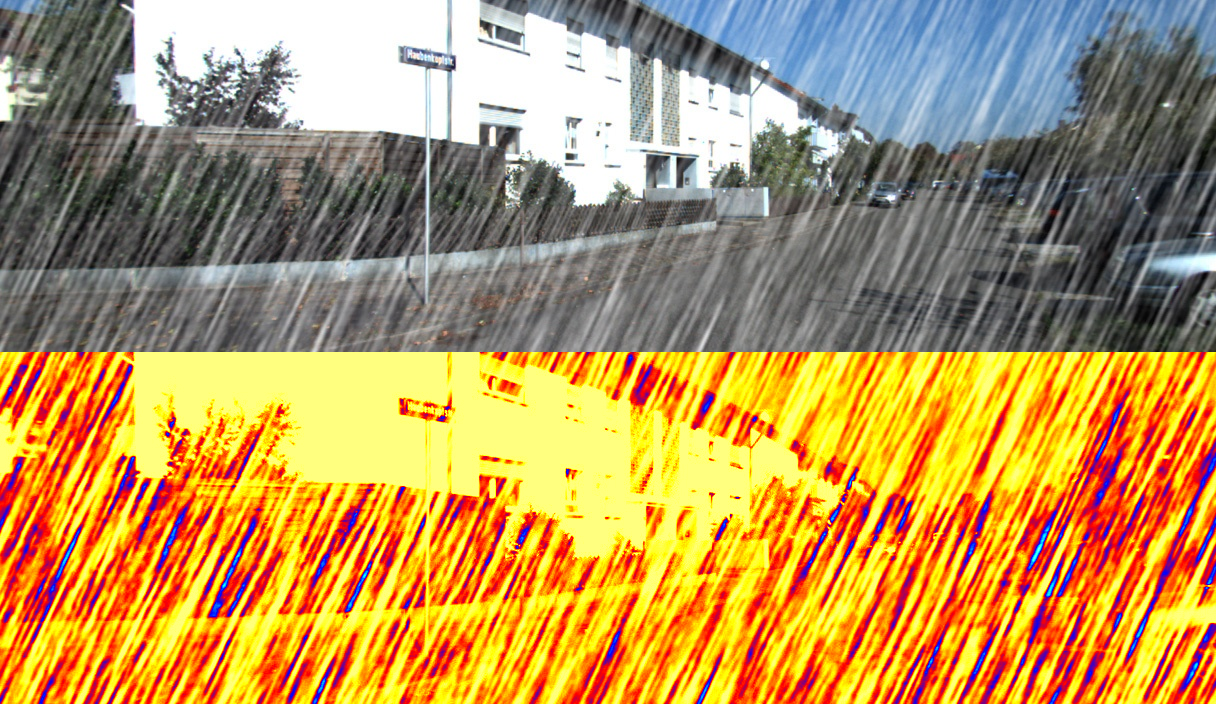}
    \includegraphics[width=1.6cm]{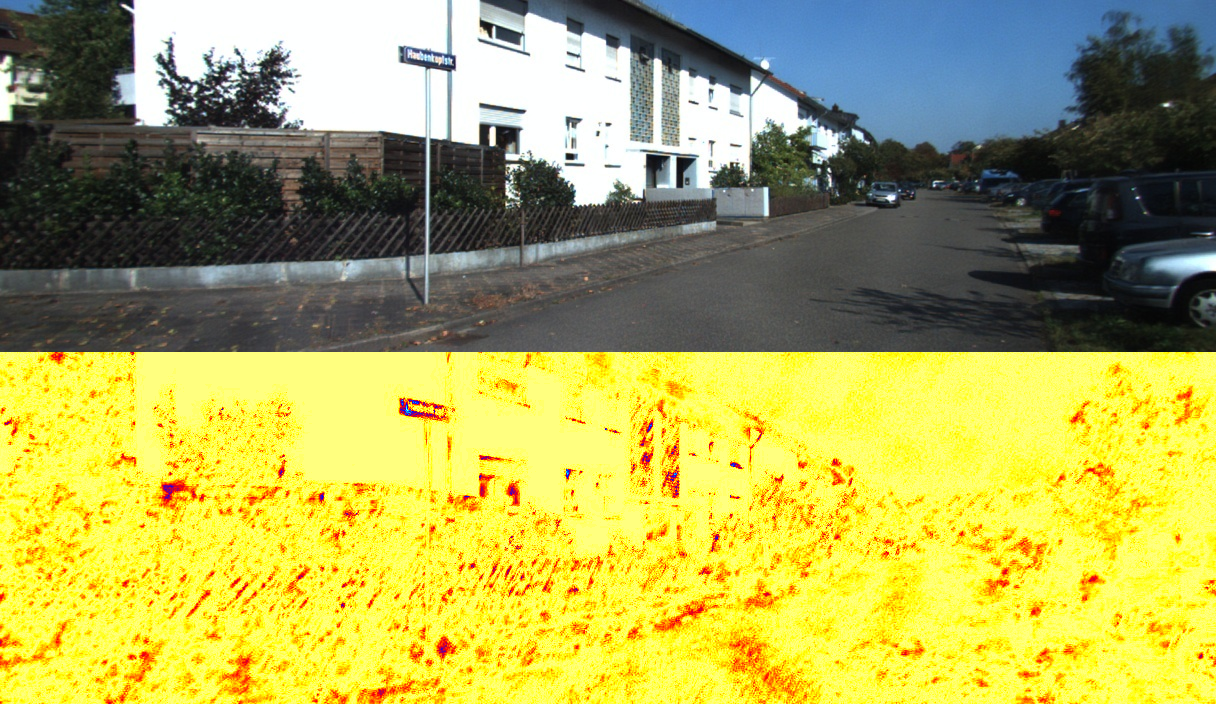}
    \includegraphics[width=1.6cm]{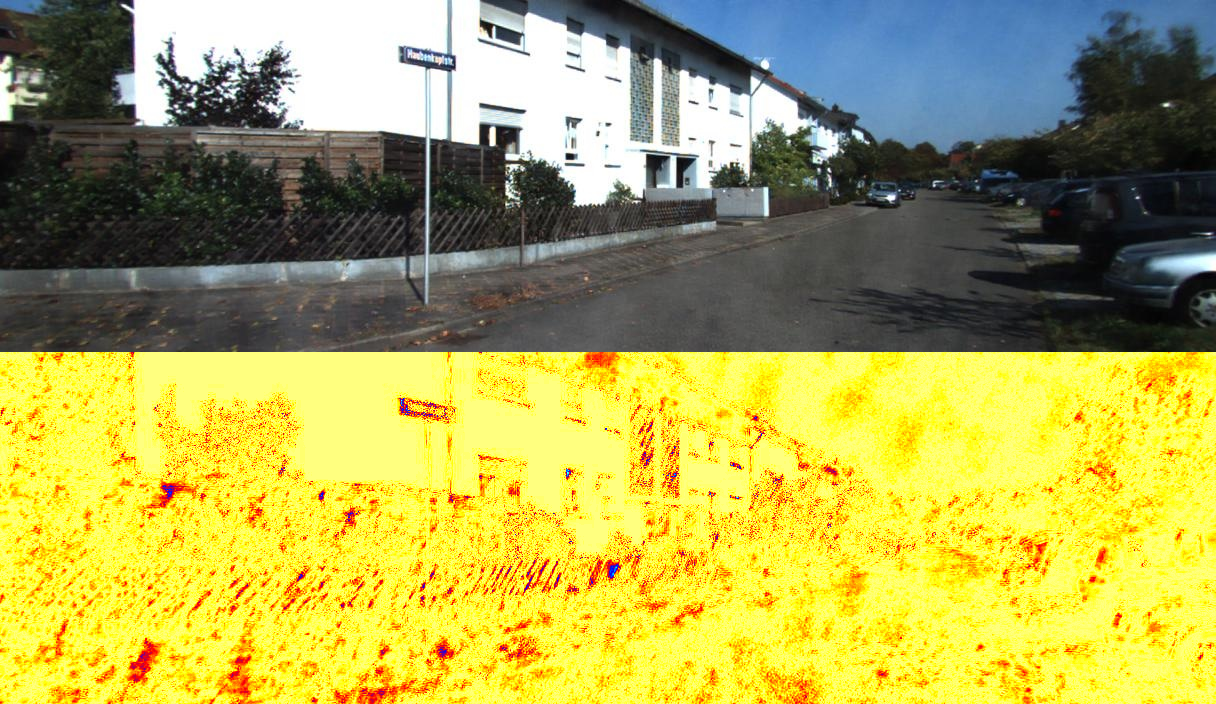}
    \includegraphics[width=1.6cm]{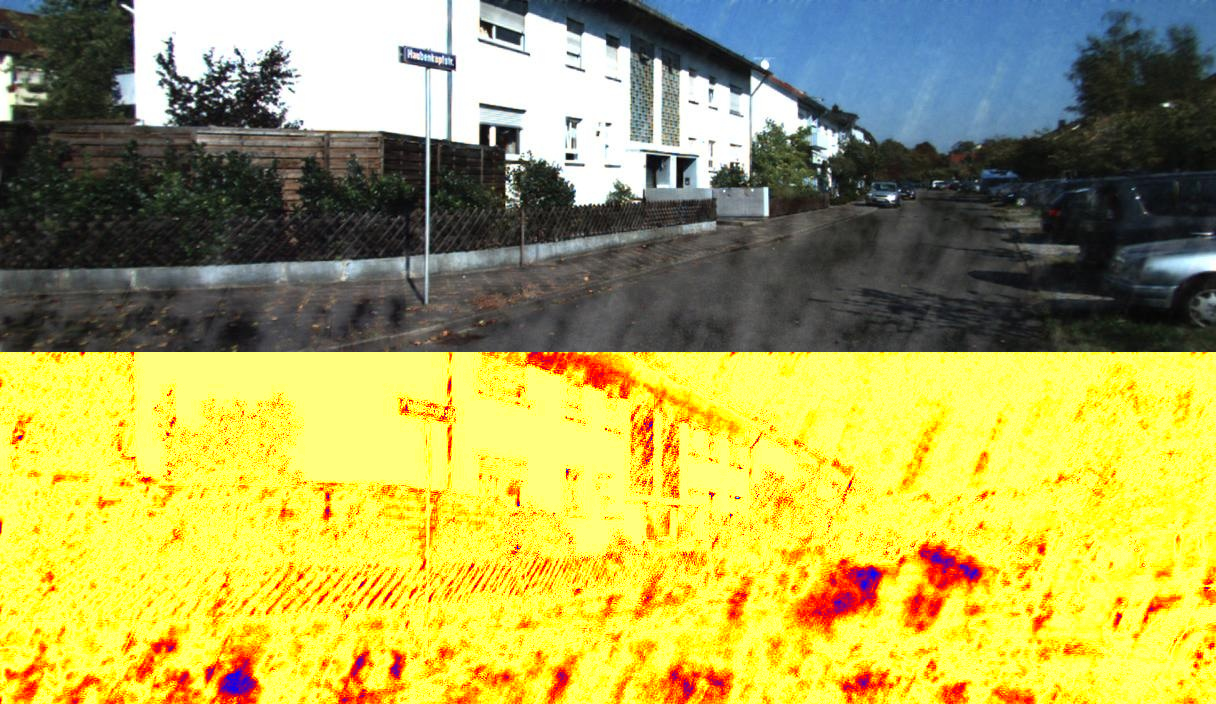}
    \includegraphics[width=1.6cm]{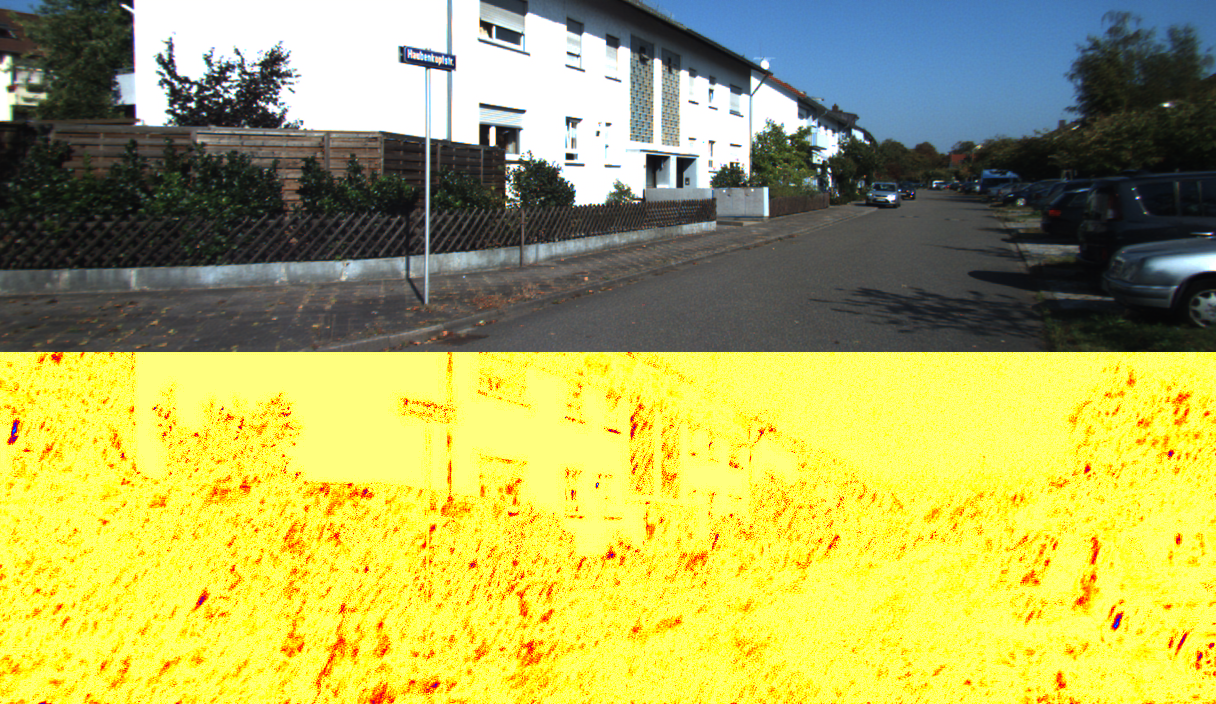}

    \vspace{0.5mm}
    \includegraphics[width=1.6cm]{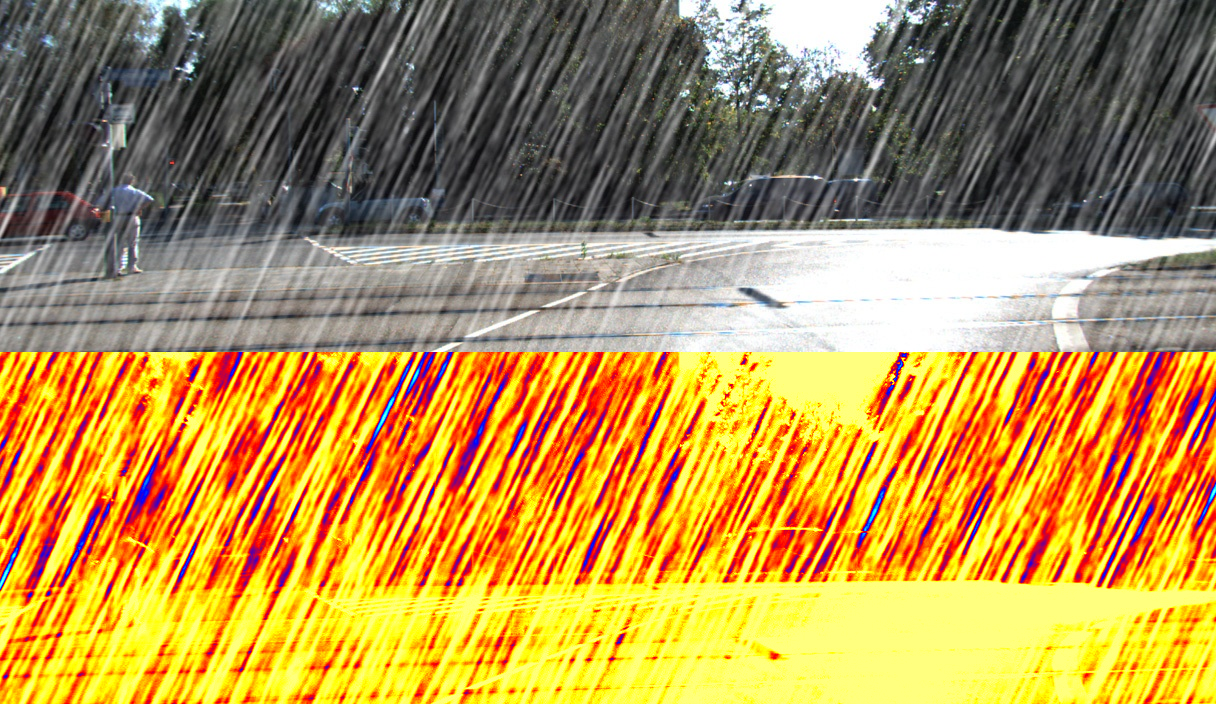}
    \includegraphics[width=1.6cm]{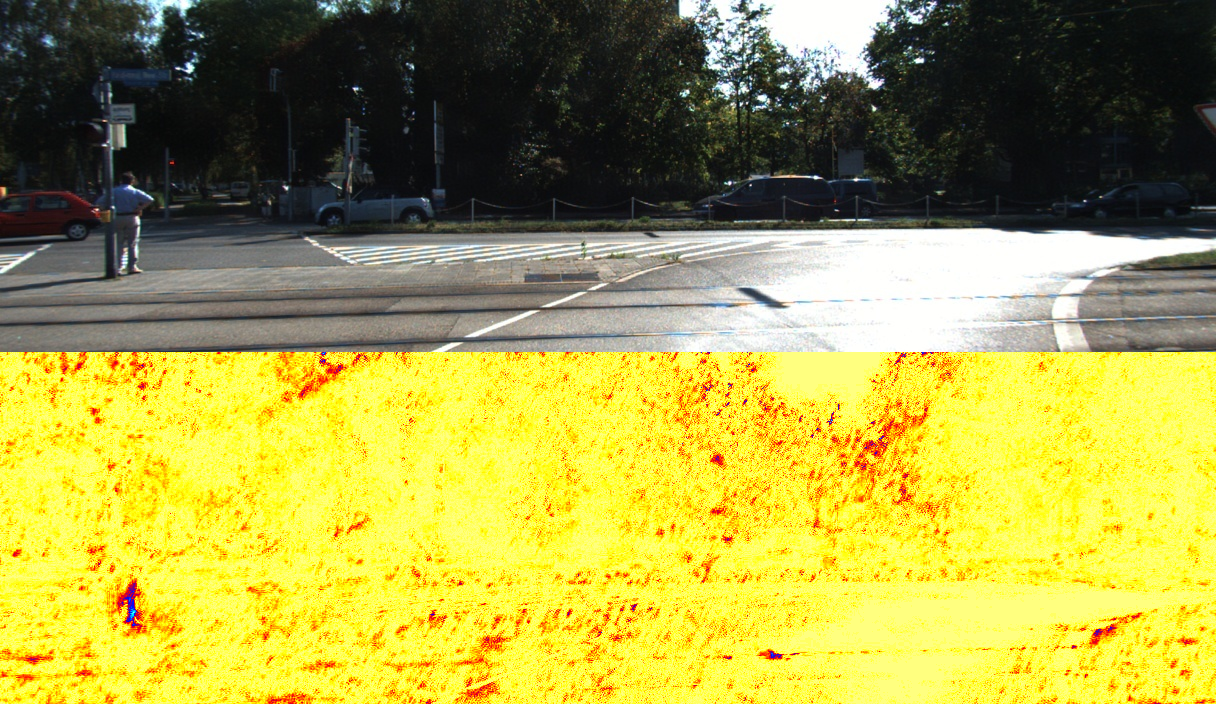}
    \includegraphics[width=1.6cm]{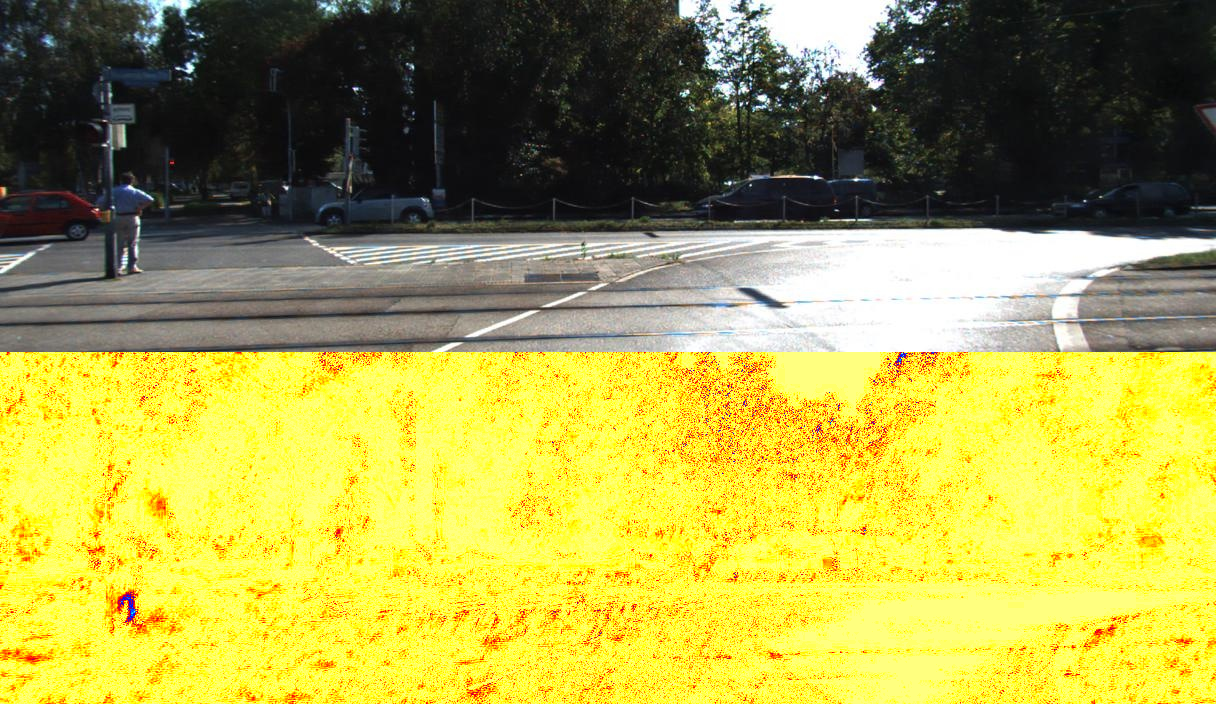}
    \includegraphics[width=1.6cm]{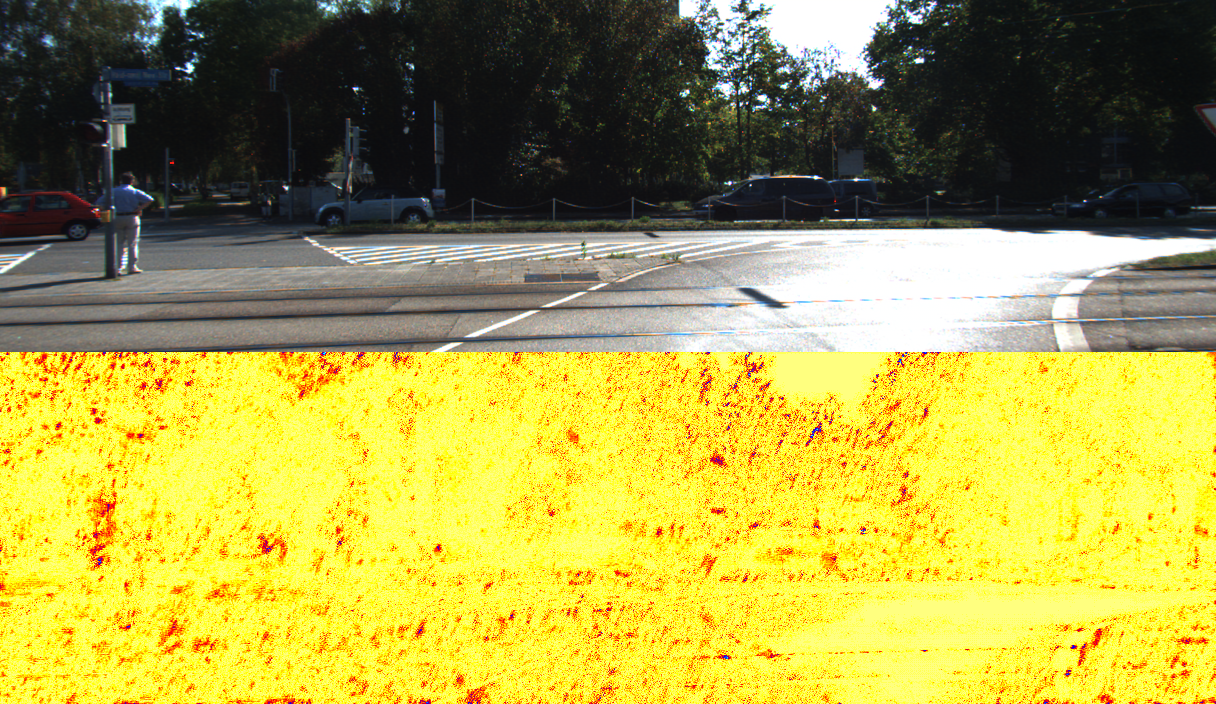}
    \includegraphics[width=1.6cm]{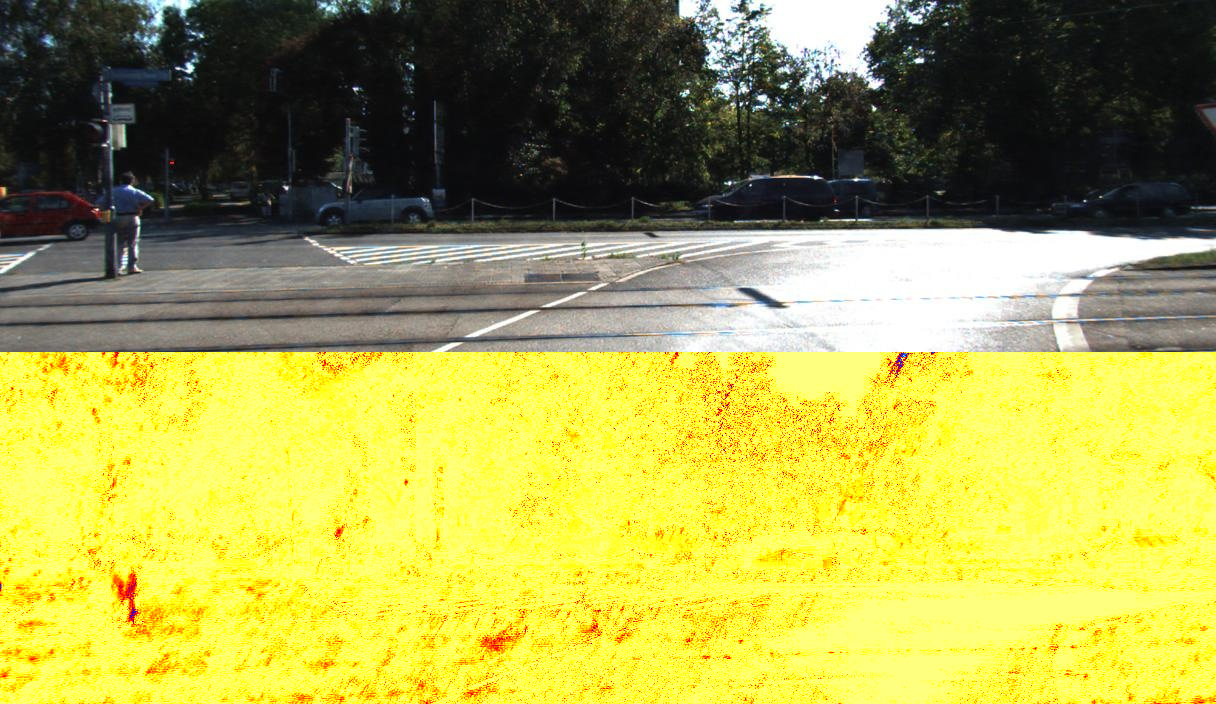}

    \vspace{0.5mm}
    \includegraphics[width=1.6cm]{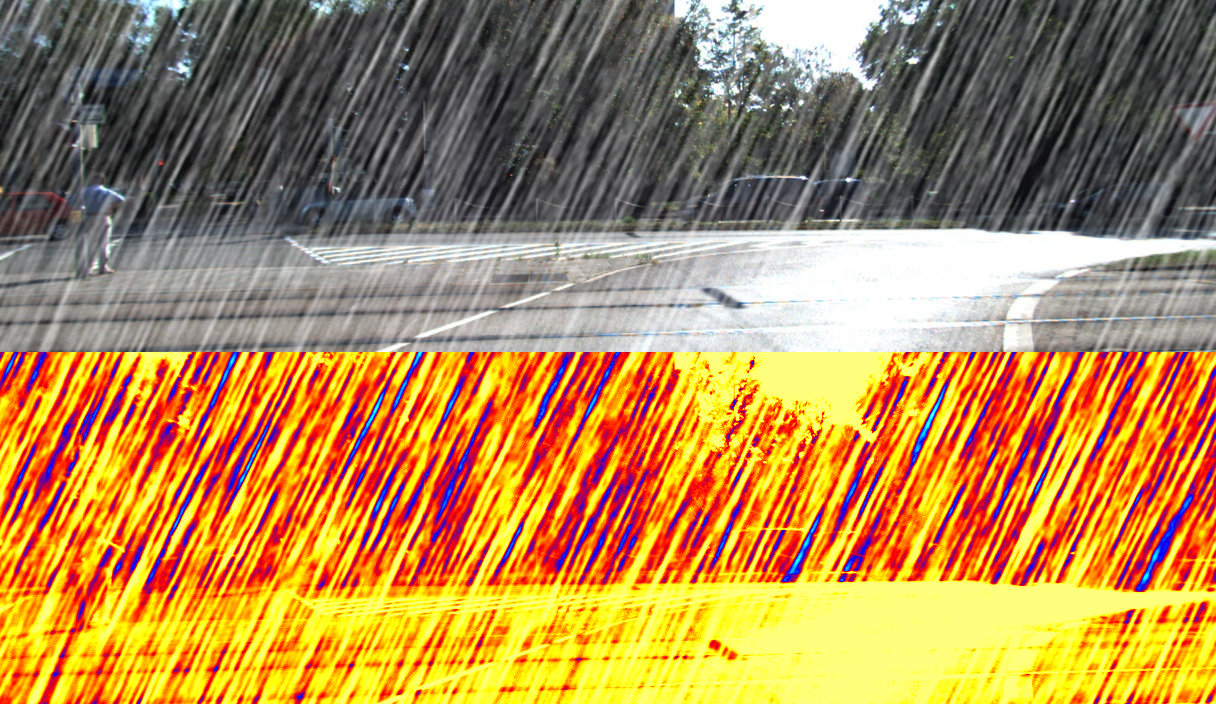}
    \includegraphics[width=1.6cm]{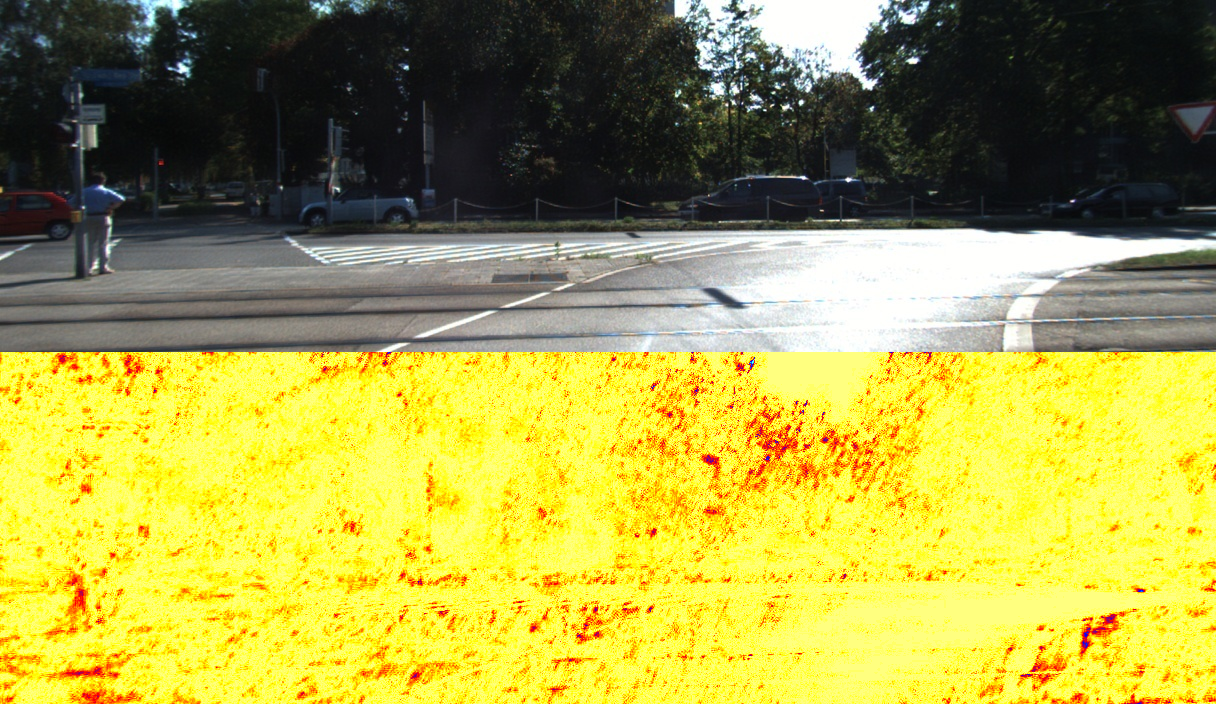}
    \includegraphics[width=1.6cm]{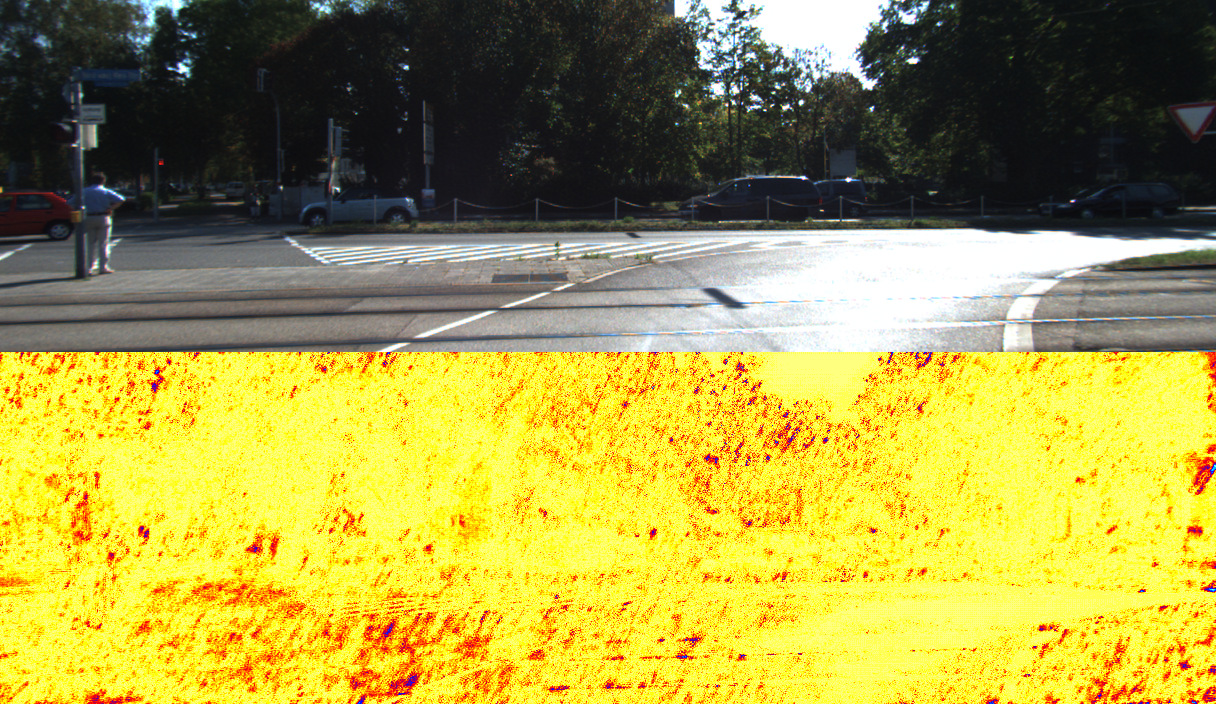}
    \includegraphics[width=1.6cm]{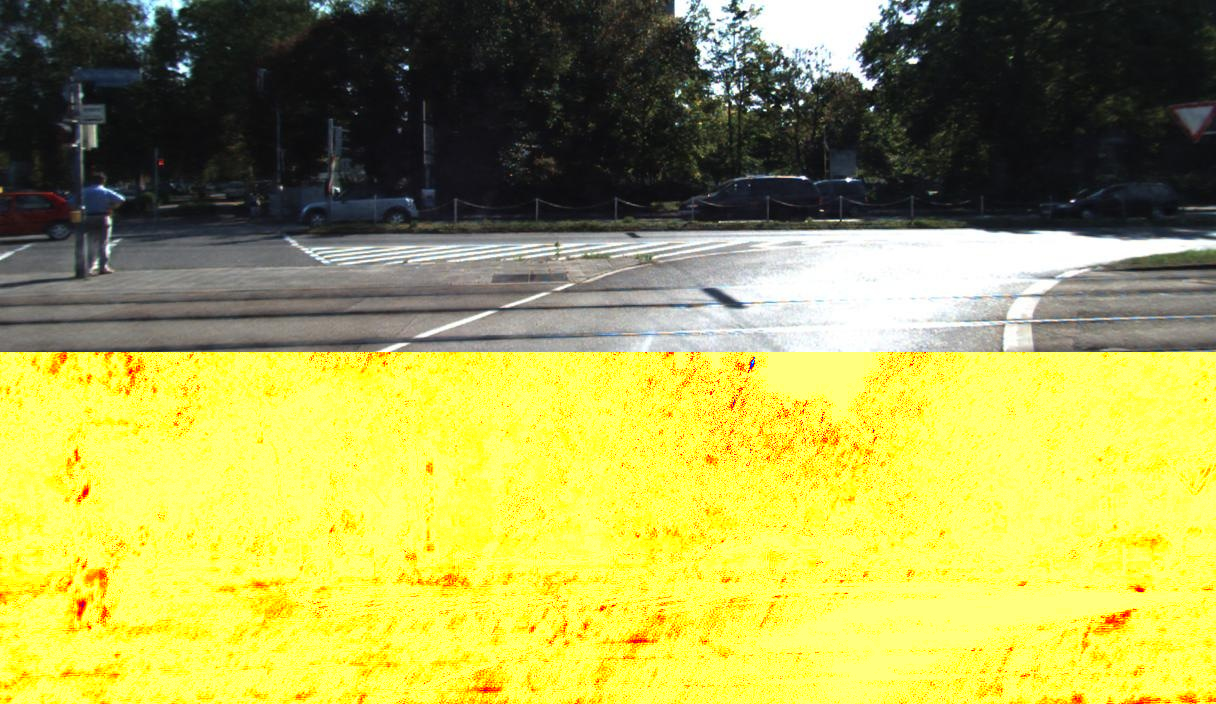}
    \includegraphics[width=1.6cm]{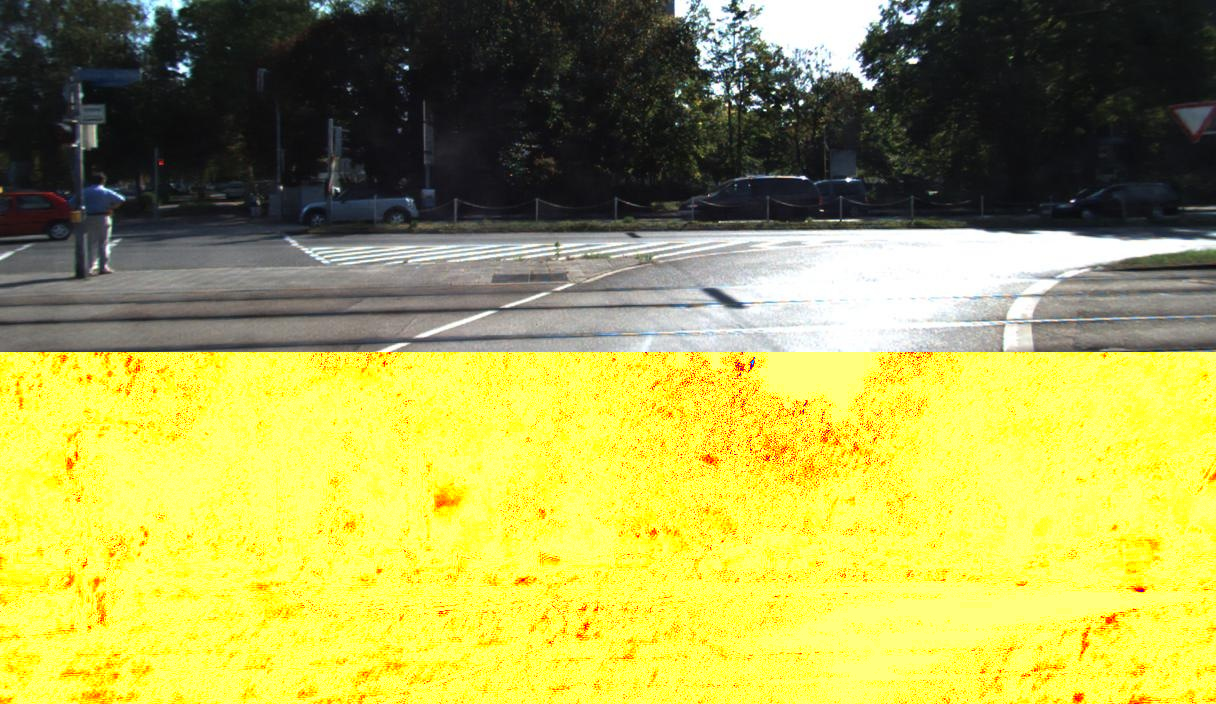}

    \vspace{0.5mm}
    \includegraphics[width=1.6cm]{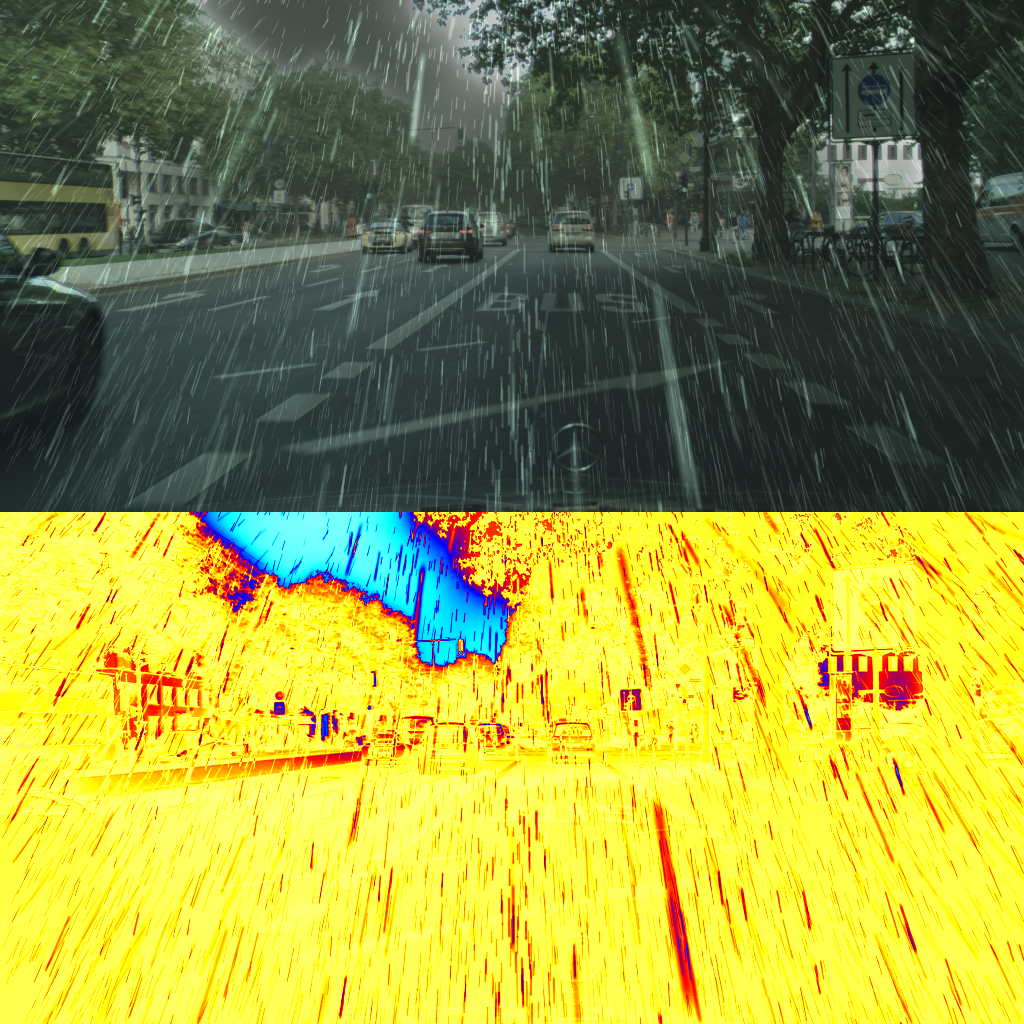}
    \includegraphics[width=1.6cm]{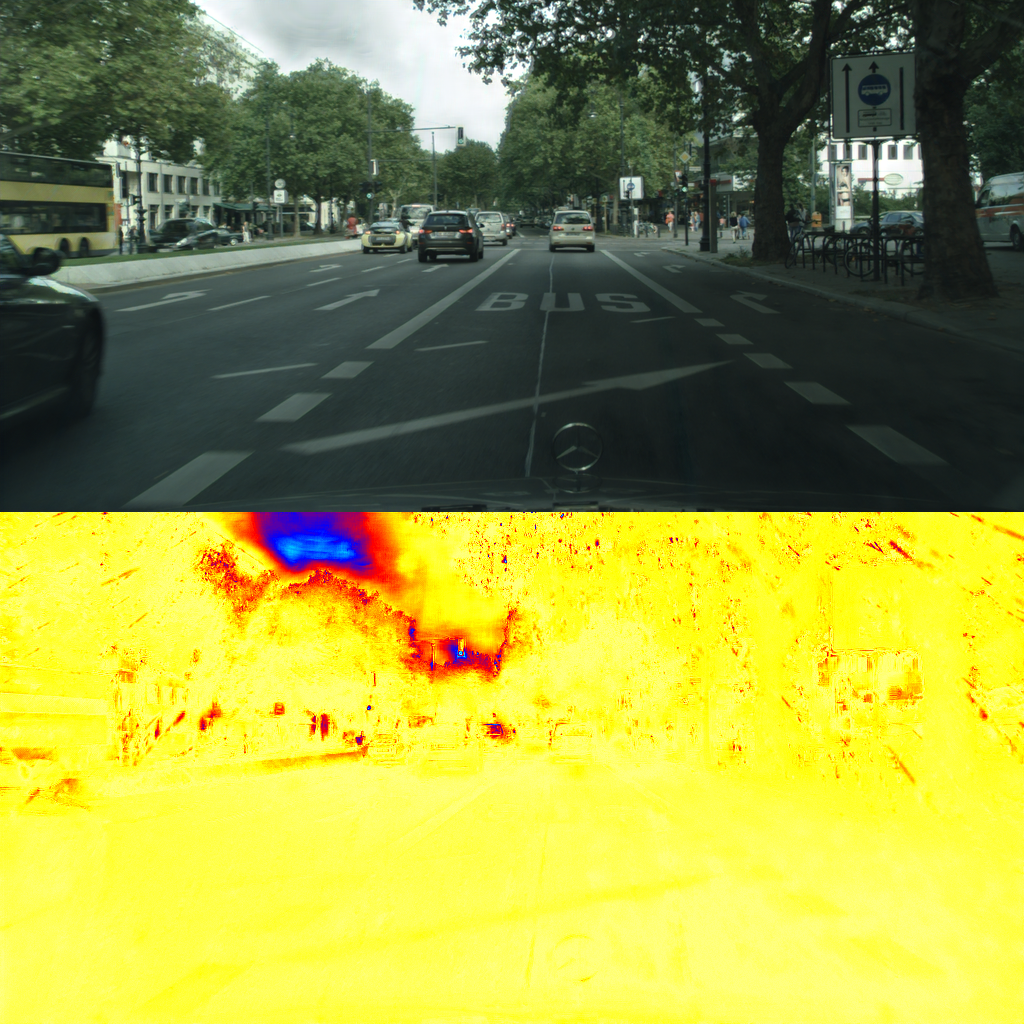}
    \includegraphics[width=1.6cm]{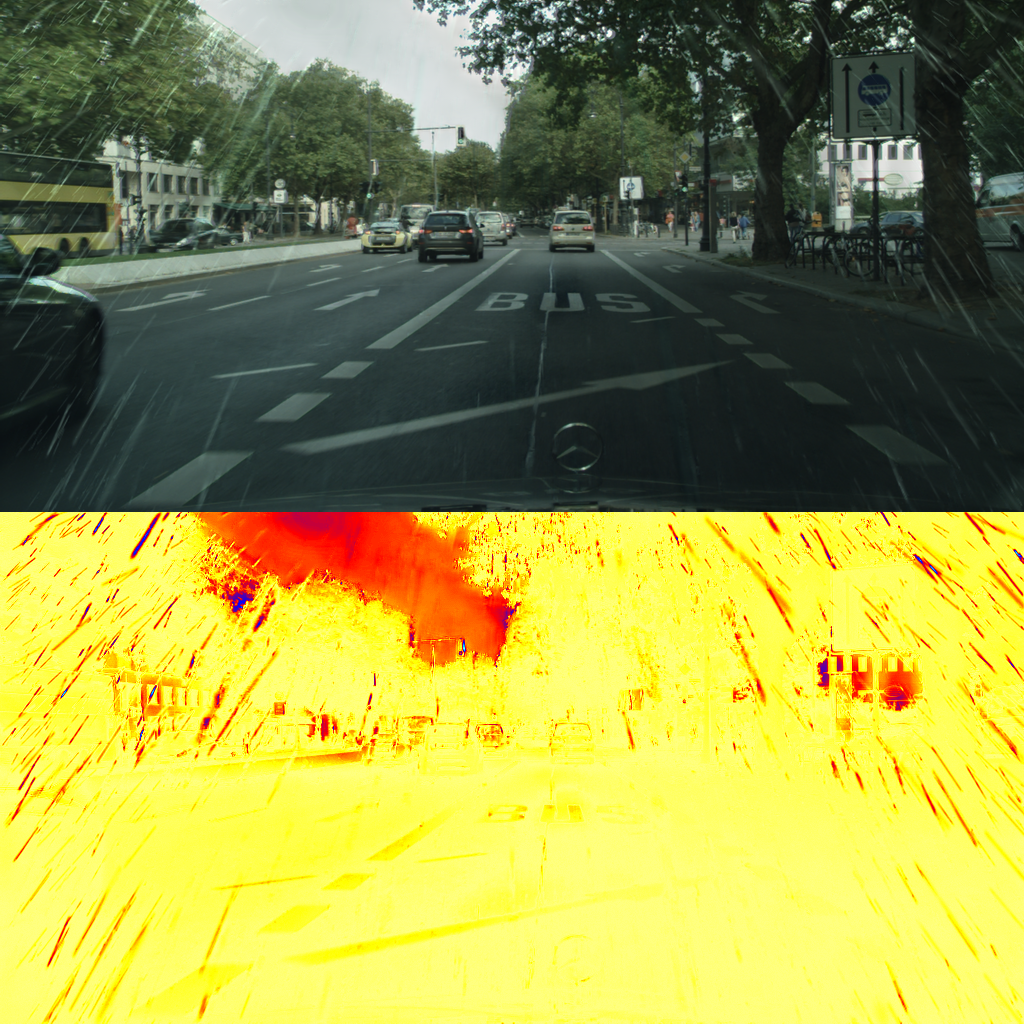}
    \includegraphics[width=1.6cm]{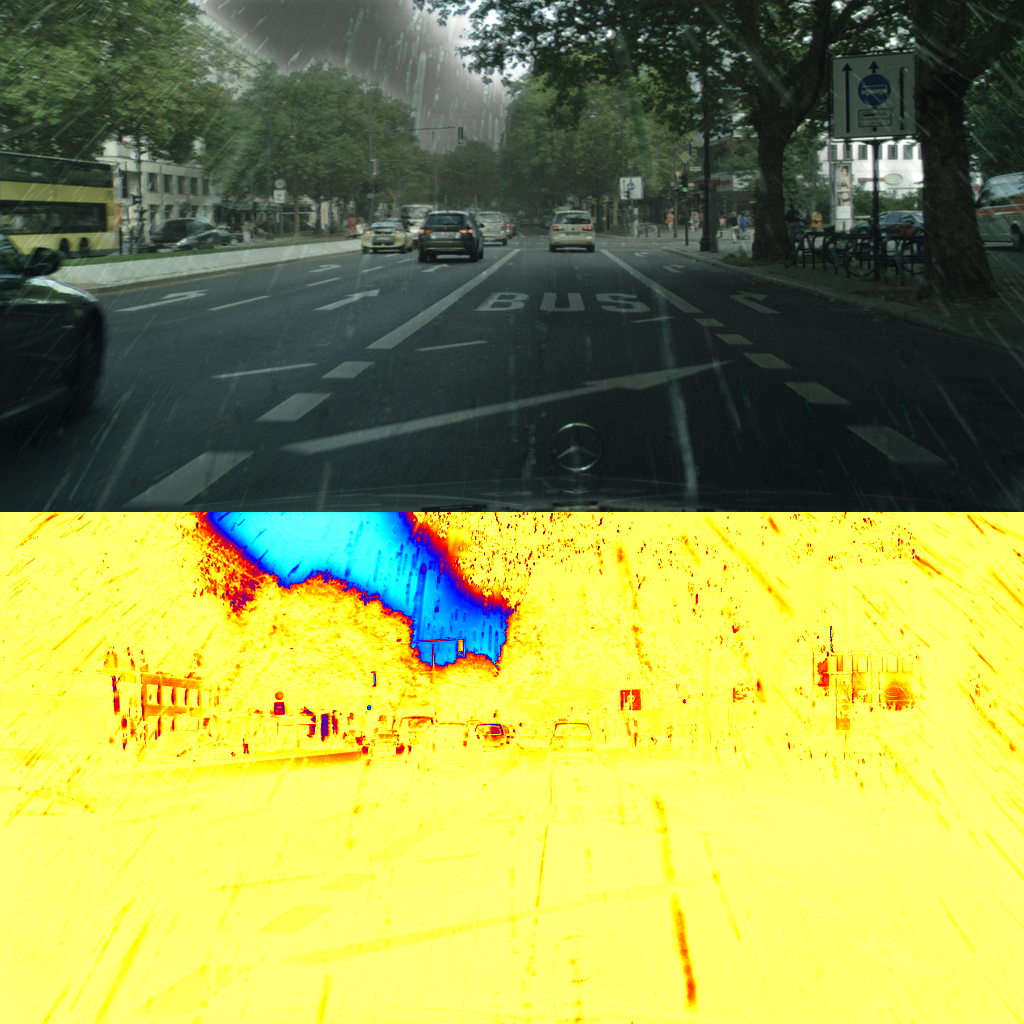}
    \includegraphics[width=1.6cm]{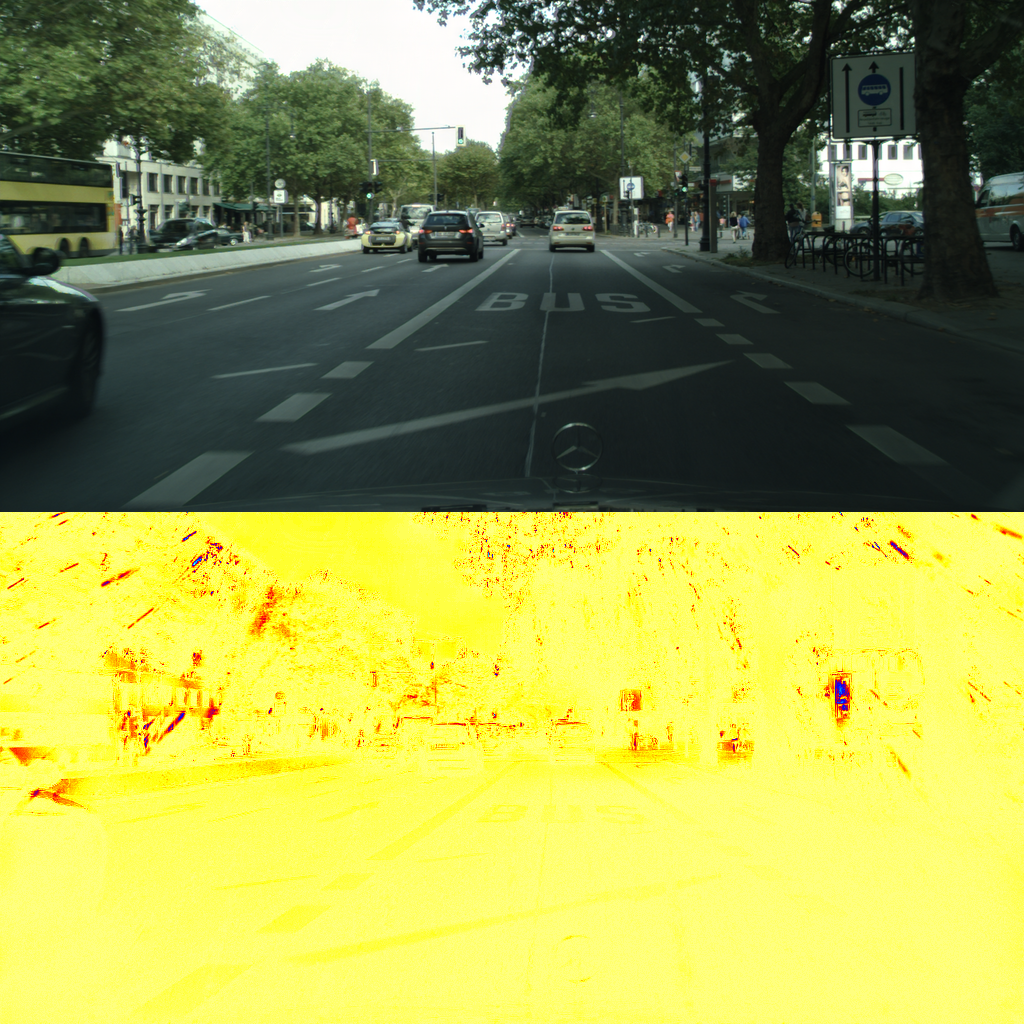}

    \vspace{0.5mm}
    \includegraphics[width=1.6cm]{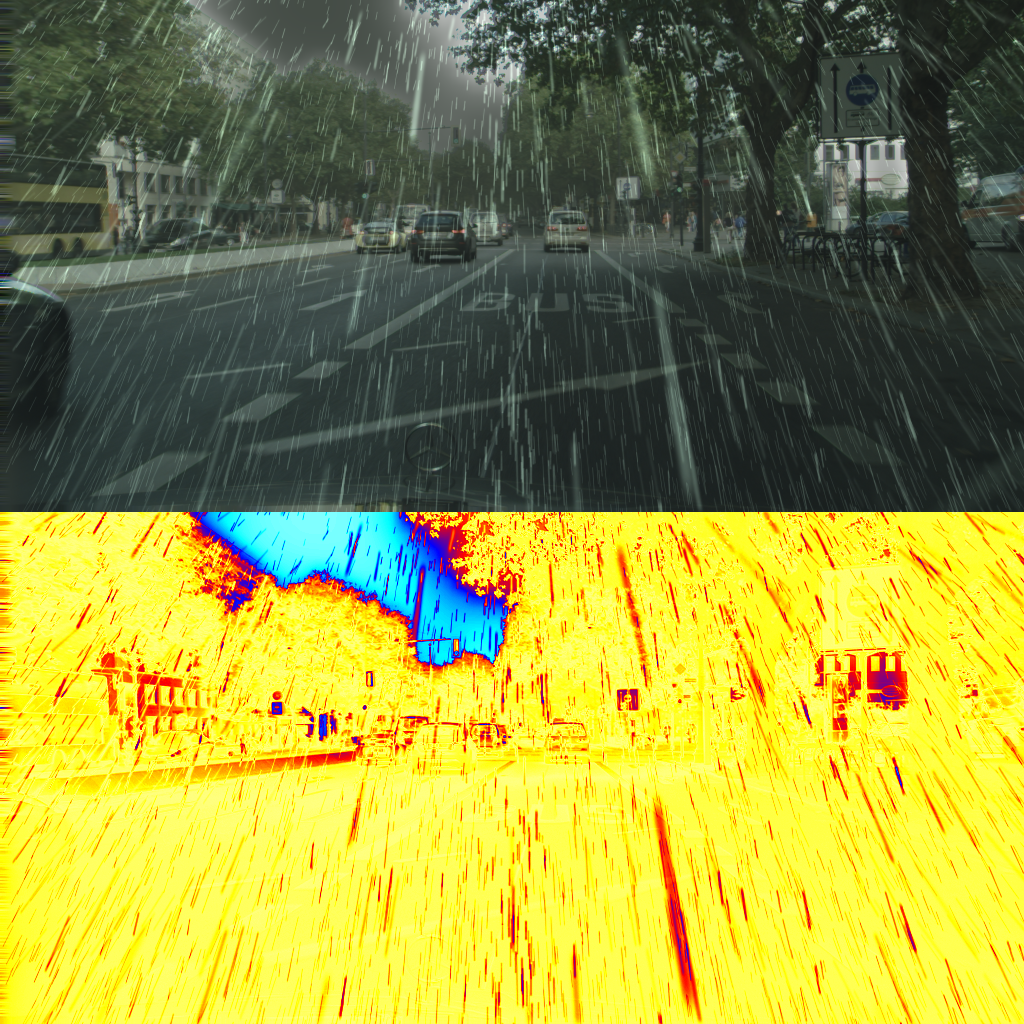}
    \includegraphics[width=1.6cm]{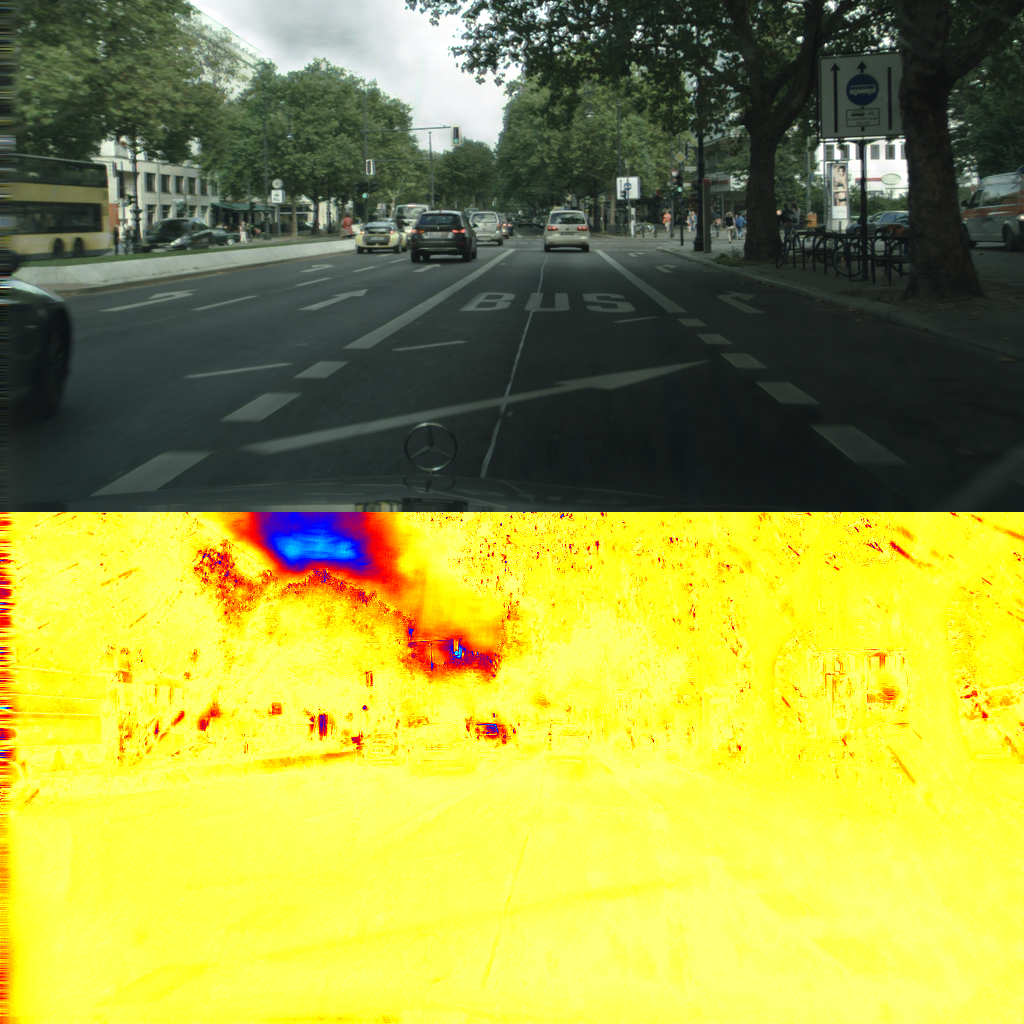}
    \includegraphics[width=1.6cm]{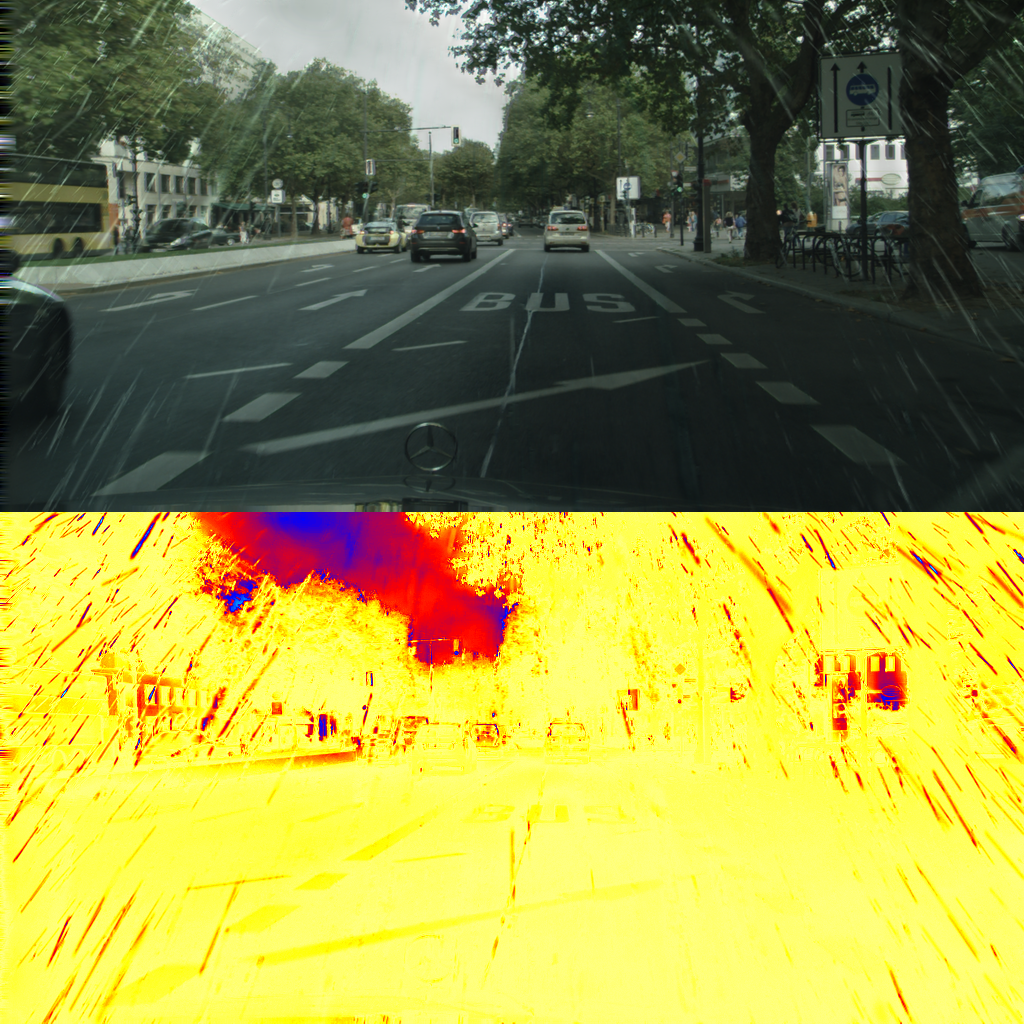}
    \includegraphics[width=1.6cm]{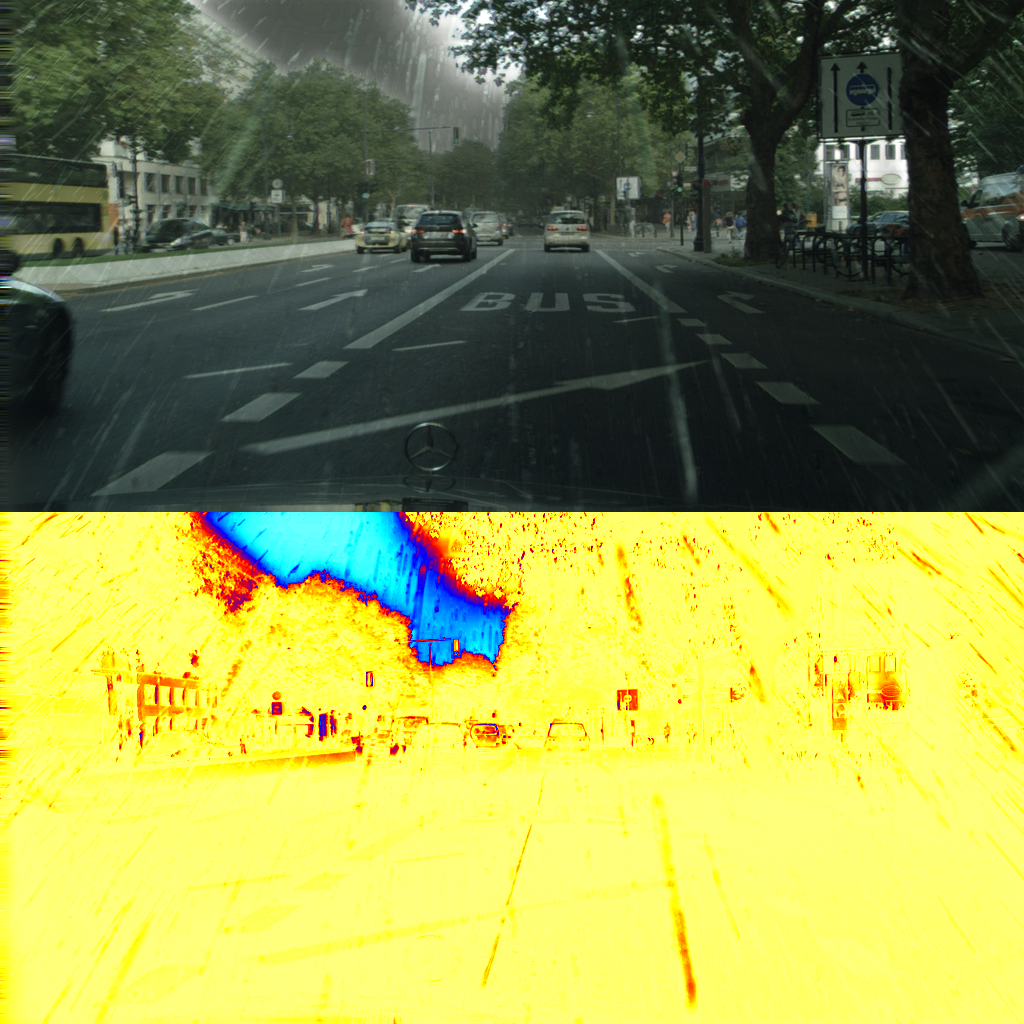}
    \includegraphics[width=1.6cm]{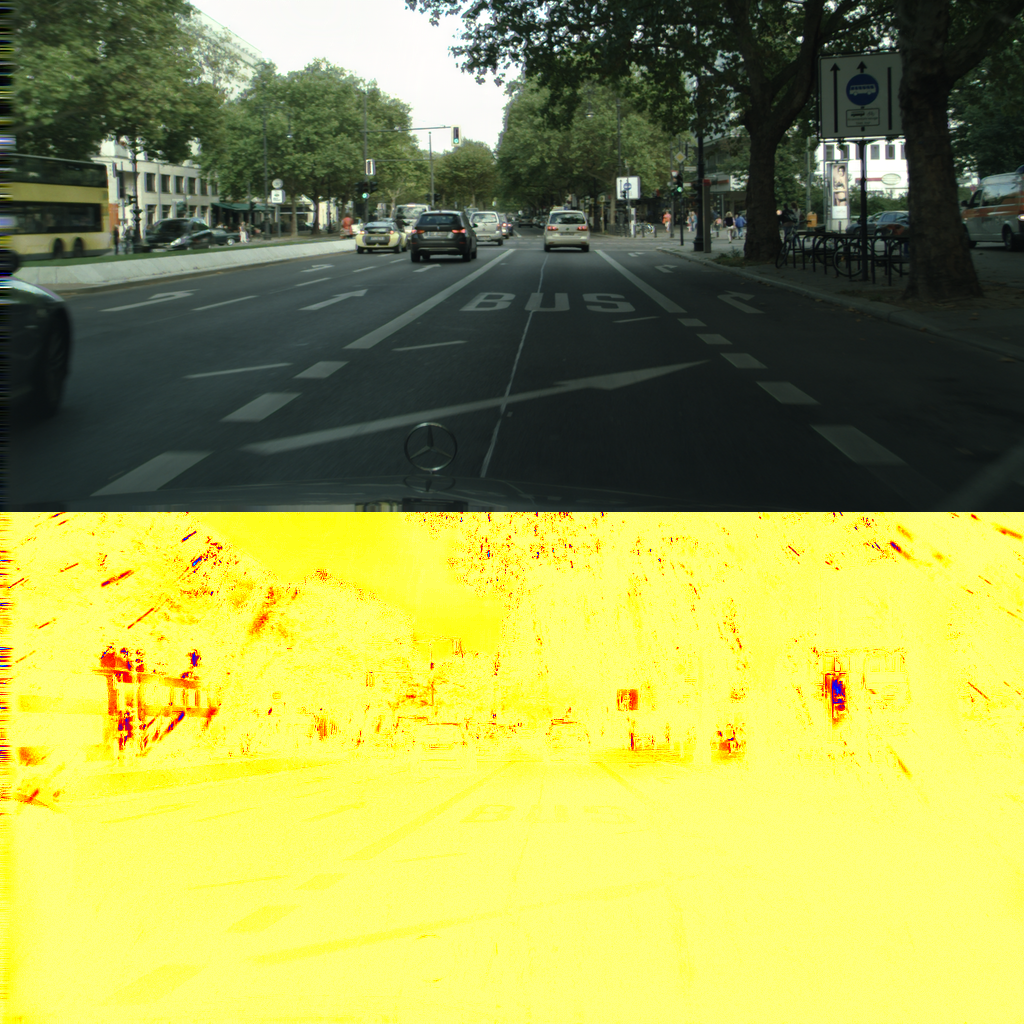}
    \caption{Visual comparisons of input images, EPRRNet \cite{zhang2022beyond}, iPASSR \cite{wang2021symmetric}, NAFSSR \cite{chu2022nafssr}, our proposed MQINet, and the corresponding error maps on RainKITTI2012 \cite{zhang2020beyond}, RainKITTI2015 \cite{zhang2020beyond} and StereoCityscapes \cite{wei2022stereo} datasets.}
    \label{fig:visual}
    \vspace{-2mm}
\end{figure}

We conduct evaluations of our MQINet alongside EPRRNet \cite{zhang2022beyond}, iPASSR \cite{wang2021symmetric}, NAFSSR \cite{chu2022nafssr} and StereoIRR \cite{wei2022stereo}.
As Table \ref{tab:metrics} detailed, our MQINet achieves the best quantitative performance across all datasets, where the scores are the average of left- and right-view sub-datasets.
We provide visual comparisons in Figure \ref{fig:visual}. Notably, when examining the error maps, our MQINet consistently produces results with the smallest errors between the results and corresponding ground truths.
For instance, when considering the bottom stereo rainy image pairs, our model eliminates rain streaks while simultaneously restoring finer texture details.

\begin{table}[H]\footnotesize
\vspace{-2mm}
    \centering
    \caption{Stereo image deraining results. Our proposed MQINet achieves the highest metrical scores on all the testing datasets.}
    \label{tab:metrics}
    \renewcommand\arraystretch{1}
    \setlength{\tabcolsep}{0.4mm}{
        \begin{tabular}{lcccccc}
            \toprule
            \multirow{2}{*}{Method} & \multicolumn{2}{c}{RainKITTI2012 \cite{zhang2022beyond}} & \multicolumn{2}{c}{RainKITTI2015 \cite{zhang2022beyond}} & \multicolumn{2}{c}{RainCityscapes \cite{wei2022stereo}} \\
            ~ & PSNR & SSIM & PSNR & SSIM & PSNR & SSIM \\
            \midrule
            iPASSR \cite{wang2021symmetric} & 35.43 & 0.975 & 35.64 & 0.973 & 28.87 & 0.963 \\
            NAFSSR \cite{chu2022nafssr} & 38.73 & 0.983 & 38.80 & 0.982 & 35.98 & 0.988 \\
            EPRRNet \cite{zhang2022beyond} & 36.18 & 0.976 & 36.35 & 0.974 & 28.07 & 0.979 \\
            StereoIRR \cite{wei2022stereo} & \blue{39.91} & \blue{0.986} & \blue{39.62} & \blue{0.984} & \blue{39.68} & \blue{0.996} \\
            \textbf{MQINet} & \red{40.36} & \red{0.991} & \red{40.28} & \red{0.989} & \red{39.83} & \red{0.997} \\
            \bottomrule
        \end{tabular}
    }
\end{table}

\subsection{Ablation Studies}

We conduct ablation experiments on RainKITTI2012 dataset \cite{zhang2022beyond} by progressively substituting the core components in NAFSSR \cite{chu2022nafssr} with our proposed CDQB, IPA, and CMIA components.
As shown in Table \ref{tab:component}, the results demonstrate that our MQINet advances Baseline by 1.63 dB and each individual contribution leads to an improvement in overall performance.

\begin{table}[H]\footnotesize
\vspace{-2mm}
     \centering
     \caption{Ablation study of individual components. Each individual contribution helps in improving the performance.}
     \label{tab:component}
     \renewcommand\arraystretch{1}
     \setlength{\tabcolsep}{2.8mm}{
     \begin{tabular}{lcccccc}
     \toprule
     \multirow{2}{*}{Model} & \multicolumn{3}{|c|}{Component} & \multirow{2}{*}{PSNR} & \multirow{2}{*}{SSIM} \\
     ~ & \multicolumn{1}{|c}{CDQB} & IPA &  \multicolumn{1}{c|}{CMIA} \\
    \midrule
    Baseline & \multicolumn{1}{|c}{~} & ~ & \multicolumn{1}{c|}{~} & 38.73 & 0.983 \\
    \romannumeral2 & \multicolumn{1}{|c}{\checkmark} & ~ & \multicolumn{1}{c|}{~} & 39.10 & 0.985 \\
    \romannumeral3 & \multicolumn{1}{|c}{\checkmark} & \checkmark & \multicolumn{1}{c|}{~} & 39.38 & 0.988 \\
    \textbf{MQINet} & \multicolumn{1}{|c}{\checkmark} & \checkmark & \multicolumn{1}{c|}{\checkmark} & \textbf{40.36} & \textbf{0.991}\\
    \bottomrule
    \end{tabular}
     }
     \vspace{-2mm}
 \end{table}

\section{Conclusion}

In this work, we introduce a multi-dimension queried and interacting network for stereo image deraining. Leveraging our proposed CDQB, IPA and CMIA, our method exhibits substantial improvements over existing methods.
Experiments highlight the efficacy of dimension-wise queries in enhancing the fundamental learning block by suppressing less-informative features. Furthermore, the incorporation of physics-aware attention results in superior shallow feature representations. Finally, multi-dimension interaction facilitates the comprehensive mutual information communication.

\end{document}